%% file: main.tex
\documentclass{article} 
\usepackage[dvipsnames]{xcolor}
\usepackage{colm_style_arxiv}

\usepackage{microtype}
\usepackage{graphicx}
\usepackage{subfigure}
\usepackage{booktabs} 
\usepackage{multirow}

\usepackage{wrapfig}
\usepackage{amsmath}
\usepackage{hyperref}
\usepackage{url}
\usepackage{breakurl}
\usepackage{bbding}  
\usepackage[capitalize,noabbrev]{cleveref}
\usepackage{colortbl}

\usepackage{xspace} 
\usepackage{enumitem} 
\usepackage{ulem} 
\usepackage{mathtools} 
\usepackage{comment}
\usepackage{annotate-equations}

\newcommand{\dataname}{\textsc{UltraInteract}\xspace}
\newcommand{\modelname}{\textsc{Eurus}\xspace}
\newcommand{\smallmodel}{\textsc{Eurus-7B}\xspace}
\newcommand{\bigmodel}{\textsc{Eurus-70B}\xspace}
\newcommand{\smallsft}{\textsc{Eurus-7B-SFT}\xspace}
\newcommand{\sft}{\textsc{Eurus-SFT}\xspace}
\newcommand{\bigsft}{\textsc{Eurus-70B-SFT}\xspace}
\newcommand{\rewardmodel}{\textsc{Eurus-RM-7B}\xspace}

\title{
Advancing LLM Reasoning Generalists with Preference Trees
}

\author{Lifan Yuan$^{1,2}$\thanks{Equal Contribution.}, 
Ganqu Cui$^{1*}$\thanks{Corresponding Authors.},
Hanbin Wang$^{3,4*}$,
Ning Ding$^1$\footnotemark[2], 
Xingyao Wang$^2$, 
Jia Deng$^5$, \\
\textbf{Boji Shan$^6$,}
\textbf{Huimin Chen$^1$,}
\textbf{Ruobing Xie$^7$,}
\textbf{Yankai Lin$^5$,}
\textbf{Zhenghao Liu$^3$,}
\textbf{Bowen Zhou$^1$,} \\
\textbf{Hao Peng$^2$,} 
\textbf{Zhiyuan Liu$^1$\footnotemark[2],}
\textbf{Maosong Sun$^1$}  \\
$^1$Tsinghua University \hfill
$^2$University of Illinois Urbana-Champaign \hfill
$^3$Northeastern University \\
$^4$ ModelBest.Inc \xspace
$^5$ Renmin University of China \xspace
$^6$ BUPT \xspace
$^7$ Tencent \\
\texttt{lifan4@illinois.edu} \hfill 
\texttt{cgq22@mails.tsinghua.edu.cn} \hfill
\texttt{wanghanbinpanda@gmail.com}
}

\NewDocumentCommand{\lifan}
{ mO{} }{\textcolor{cyan}{\textsuperscript{\textit{Lifan}}\textsf{\textbf{\small[#1]}}}}
\NewDocumentCommand{\hao}
{ mO{} }{\textcolor{purple}{\textsuperscript{\textit{Hao}}\textsf{\textbf{\small[#1]}}}}
\NewDocumentCommand{\xingyao}
{ mO{} }{\textcolor{magenta}{\textsuperscript{\textit{Xingyao}}\textsf{\textbf{\small[#1]}}}}

\usepackage{hyperref}
\hypersetup{
	colorlinks=true,
	linkcolor=cyan,
	filecolor=blue,      
	urlcolor=red,
	citecolor=cyan,
}

\NewDocumentCommand{\dn}
{ mO{} }{\textcolor{red}{\textsuperscript{\textit{dn}}\textsf{\textbf{\small[#1]}}}}
\NewDocumentCommand{\gq}
{ mO{} }{\textcolor{brown}{\textsuperscript{\textit{ganqu}}\textsf{\textbf{\small[#1]}}}}
\NewDocumentCommand{\hanbin}
{ mO{} }{\textcolor{orange}{\textsuperscript{\textit{hanbin}}\textsf{\textbf{\small[#1]}}}}
\colmfinalcopy 
\begin{document}

\maketitle

\begin{abstract} 
\looseness=-1
We introduce \textbf{\modelname}, a suite of large language models (LLMs) optimized for reasoning.
Finetuned from Mistral-7B and CodeLlama-70B, \modelname models achieve state-of-the-art results among open-source models on a diverse set of benchmarks covering mathematics, code generation, and logical reasoning problems. 
Notably, \modelname-70B beats GPT-3.5 Turbo in reasoning through a comprehensive benchmarking across 12 tests covering five tasks, and achieves a 33.3\% pass@1 accuracy on LeetCode and 32.6\% on TheoremQA, two challenging benchmarks, substantially outperforming existing open-source models by margins more than 13.3\%.
The strong performance of \modelname can be primarily attributed to \textbf{\dataname}, our newly-curated large-scale, high-quality alignment dataset specifically designed for complex reasoning tasks. 
\dataname can be used in both supervised fine-tuning and preference learning.
For each instruction, it includes a preference tree consisting of
(1) reasoning chains with diverse planning strategies in a unified format, 
(2) multi-turn interaction trajectories with the environment and the critique,
and (3) pairwise data to facilitate preference learning. 
\dataname allows us to conduct an in-depth exploration of preference learning for reasoning tasks.
Our investigation reveals that some well-established preference learning algorithms may be less suitable for reasoning tasks compared to their effectiveness in general conversations.
Inspired by this, we derive a novel reward modeling objective which, together with \dataname, leads to a strong reward model.\footnote{Models and data are available at: \url{https://github.com/OpenBMB/Eurus}. }

\end{abstract}

\input{sections/1_introduction}

\input{sections/3_method}
\input{sections/_model_intro}
\input{sections/4_experiment}

\input{sections/2_related_work}
\input{sections/6_conclusion}

\bibliography{main}
\bibliographystyle{colm2024_conference}

\newpage
\appendix

\input{sections/7_appendix}

\end{document}

%% file: sections/1_introduction.tex
\begin{figure}[htb]
    \vspace{-25pt}
    \centering
    \includegraphics[width=.6\linewidth]{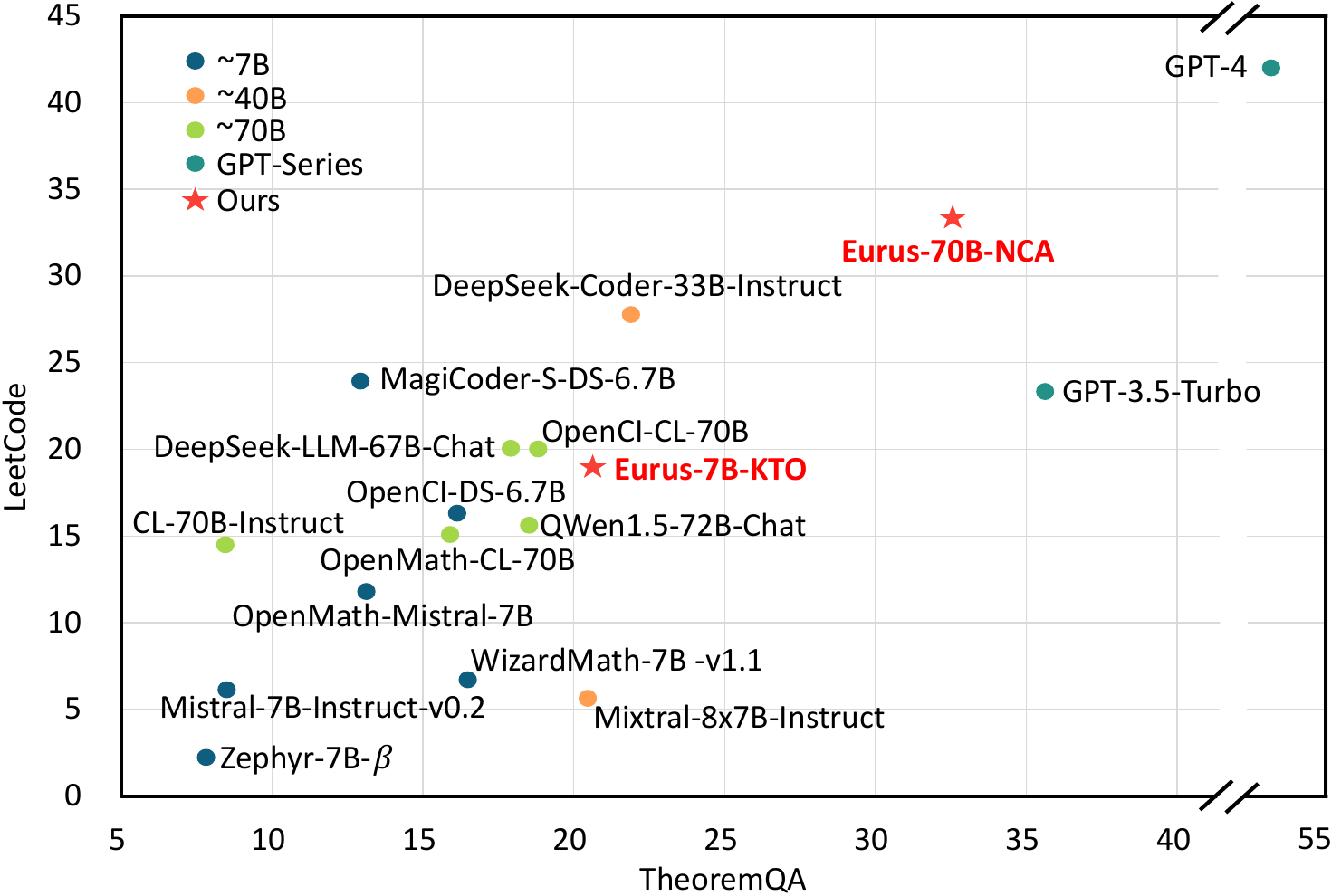}
    \vspace{-10pt}
    \caption{Evaluation results on LeetCode and TheoremQA, two challenging OOD coding and math benchmarks with only test sets. Our \smallmodel is comparable with baselines that are 10x larger and \bigmodel is the only one on par with GPT-3.5 Turbo.
    }
    \label{fig:leetcode_theoremqa}
    \vspace{-20pt}
\end{figure}
\section{Introduction}
Current alignment techniques have significantly advanced the development of open-source large language models (LLMs) that effectively meet user expectations and align with human values~\citep{Touvron2023Llama2O,tunstall2023zephyr}.
On complex reasoning, success has been achieved by specializing models for specific capabilities, such as coding~\citep{magicoder,deepseek-coder,  zheng2024opencodeinterpreter} and solving math problems~\citep{fu23specializing,yue2023mammoth, luo2023wizardmath, toshniwal2024openmath}. 
However, these models still fall short, by large margins, of the most advanced proprietary models in their all-around capabilities to tackle a diverse range of challenging problems.
We conjecture that this performance gap can be primarily attributed to (1) the lack of high-quality alignment data and (2) the underexploration of preference learning techniques for improving models' complex reasoning capabilities.
In this paper, we take strides towards bridging this gap by addressing both factors and developing \modelname.

\modelname consists of a suite of LLMs finetuned from Mistral-7B~\citep{jiang2023mistral} and CodeLLaMA-70B~\citep{roziere2023code}.
Across a diverse set of complex reasoning benchmarks that are mostly out-of-distribution (OOD),
\modelname achieves state-of-the-art overall performance among all open-source models.
In particular, \modelname excels in solving challenging problems that often require sophisticated planning, reasoning, tool integration, and the ability to interact with and learn from the environment and users. 
As shown in Figure~\ref{fig:leetcode_theoremqa}, on university-level STEM questions TheoremQA~\citep{Chen2023TheoremQAAT} and competition-level coding problems LeetCode Contest~\citep{deepseek-coder}, \modelname-70B significantly outperforms all open-source models, achieving comparable performance to GPT-3.5 Turbo.

\looseness=-1
\modelname models are trained on \dataname, our newly-curated, large-scale, and high-quality alignment data specifically designed to improve LLMs' reasoning capabilities.
\dataname consists of a diverse set of instructions spanning math, coding, and logical reasoning problems from 12 established datasets.
For each instruction, \dataname collects a preference tree that includes:
(1) \textbf{Diverse planning strategies in a unified pattern,}  such as sequential processing \citep{Wei2022ChainOT} and tool creation \citep{Qian2023CREATORTC}, followed by executing step-by-step actions formatted in either text or code, to provide divserse reasoning trajectories.
(2) \textbf{Multi-turn interaction trajectories with the environment and the critique,} to improve models' capabilities to learn from feedback and correct previous errors~\citep{Wang2023MINTEL}.
(3) \textbf{Paired correct and incorrect actions organized in tree structures,} to facilitate preference learning. 
In total, \dataname contains 86K instructions and 220K action pairs, where each pair consists of an instruction, a correct response, and an incorrect one. 
Conceptually, \dataname's data resemble imbalanced binary trees as shown in Figure \ref{fig:tree}.

\dataname can be used in both supervised fine-tuning and preference learning.
Our experiments show that, using \dataname along with established datasets in instruction fine-tuning already achieves strong performance.
\dataname further facilitates preference learning for reasoning tasks, 
improving the performance even further
with KTO \citep{Ethayarajh2024KTOMA} and NCA \citep{Chen2024NoiseCA}. 
Surprisingly, applied to an instruction finetuned \modelname model, DPO~\citep{Rafailov2023DirectPO} hurts the performance.  

\looseness=-1
Through careful analysis, we provide evidence that the performance in reasoning correlates with the value of rewards of chosen data---a higher final reward often indicates a better reasoning capability. 
Besides, our investigation suggests that DPO may be less suitable for reasoning tasks than KTO and NCA. 
Inspired by this fresh finding, we devise a new objective for reward modeling to augment the Bradley-Terry objective \citep{Bradley1952RankAO}, explicitly encouraging training to increase the absolute rewards of chosen solution and decrease those of rejected data.
Furthermore, \dataname leads to our reward model \rewardmodel, which achieves a better correlation with human annotators than all existing models on 
AutoJ~\citep{Li2023GenerativeJF} and MT-Bench~\citep{Zheng2023JudgingLW}, including GPT-4~\citep{openai2023gpt4}. \rewardmodel demonstrates especially strong preference modeling performance on reasoning tasks. 


Checkpoints of our \modelname models, accompanying \dataname alignment data to reproduce this research, will be publicly available.

%% file: sections/3_method.tex
\section{\dataname: Tree-structured Alignment Data for Reasoning}

\begin{wrapfigure}{r}{0.65\textwidth}
    \centering
    \vspace{-15pt}
    \includegraphics[width=\linewidth]{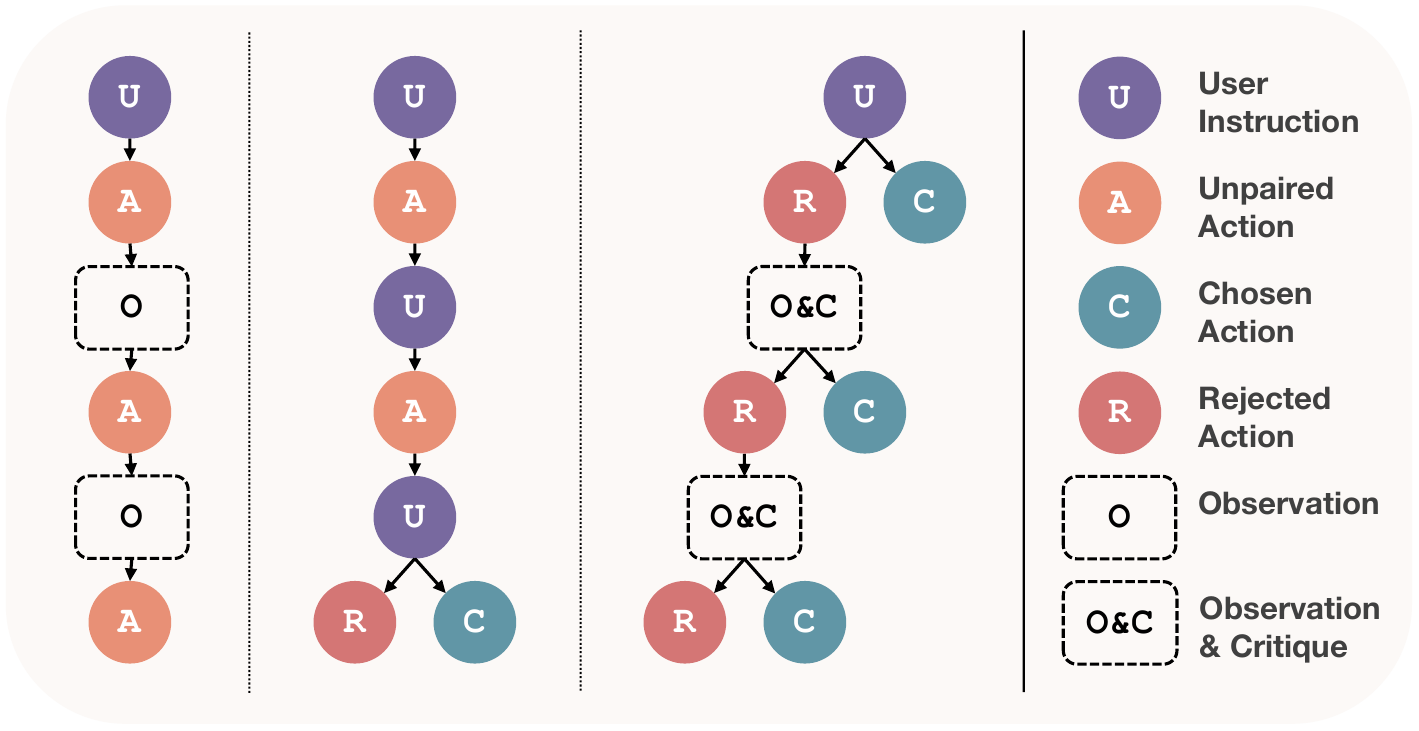}
    \vspace{-10pt}
    \caption{Left: CodeActInstruct \citep{Wang2024ExecutableCA} and Code-Feedback \citep{zheng2024opencodeinterpreter}; Middle: HH-RLHF \citep{Bai2022TrainingAH}; Right: \dataname. Each instruction in \dataname is constructed as a preference tree. 
    }
    \label{fig:tree}
    \vspace{-8pt}
\end{wrapfigure}

\looseness=-1
Solving complex problems often requires the model's capability in planning and reasoning, integrating with tools, and interacting with and learning from both the environment and the users.
This is reflected in \dataname's design choices:
(1) Its instructions are diverse, challenging, and of a large scale (\S\ref{sec:instruction_selection}); 
(2) It provides multi-turn trajectories that solve the input instruction through multiple turns of interaction with and learning from the environment and critique.
At each turn, it breaks down the problem into smaller ones (\S\ref{sec:trajectory_generation}).
(3) \dataname includes pairwise data to facilitate preference learning (\S\ref{sec:preference_data}).


Conceptually, \dataname collects a preference tree for each instruction, with the instruction being the root and each action a node (Figure \ref{fig:tree}).
A trajectory is a root-to-leaf path consisting of a sequence of actions.
In each preference tree, all nodes of correct actions and all trajectories ending with correct actions can be used for SFT. Paired correct and incorrect nodes or trajectories can be used for preference learning.

\subsection{Instruction Selection Emphasizing Complexity, Quality, and Diversity}
\label{sec:instruction_selection}
We target three representative reasoning tasks: math problem-solving, code generation, and logical reasoning. 
The complexity, quality, and diversity of the alignment data are crucial to the model's performance \citep{Liu2023WhatMG}.
Following \citet{Wang2023MINTEL}, we select challenging problems that GPT-3.5-Turbo fails to solve. 
We intentionally restrict the selection of the datasets to those with ground-truth solutions, aiming to ensure high-quality oversight signals rather than relying on LLM-as-a-judge annotation \citep{weyssow2024codeultrafeedback}. 
Besides, the gold solutions also serve as references for the critique model to generate feedback. 
To promote \dataname's diversity, we pick datasets of different categories. 
For each dataset, we include distinct reasoning patterns based on question categories or formulations necessary to solve the problems.
Table~\ref{tab:data_list} summarizes the datasets selected by \dataname. 
Except for MATH, none of the training datasets is used in our evaluation.

\begin{figure*}
    \centering
    \includegraphics[width=\linewidth]{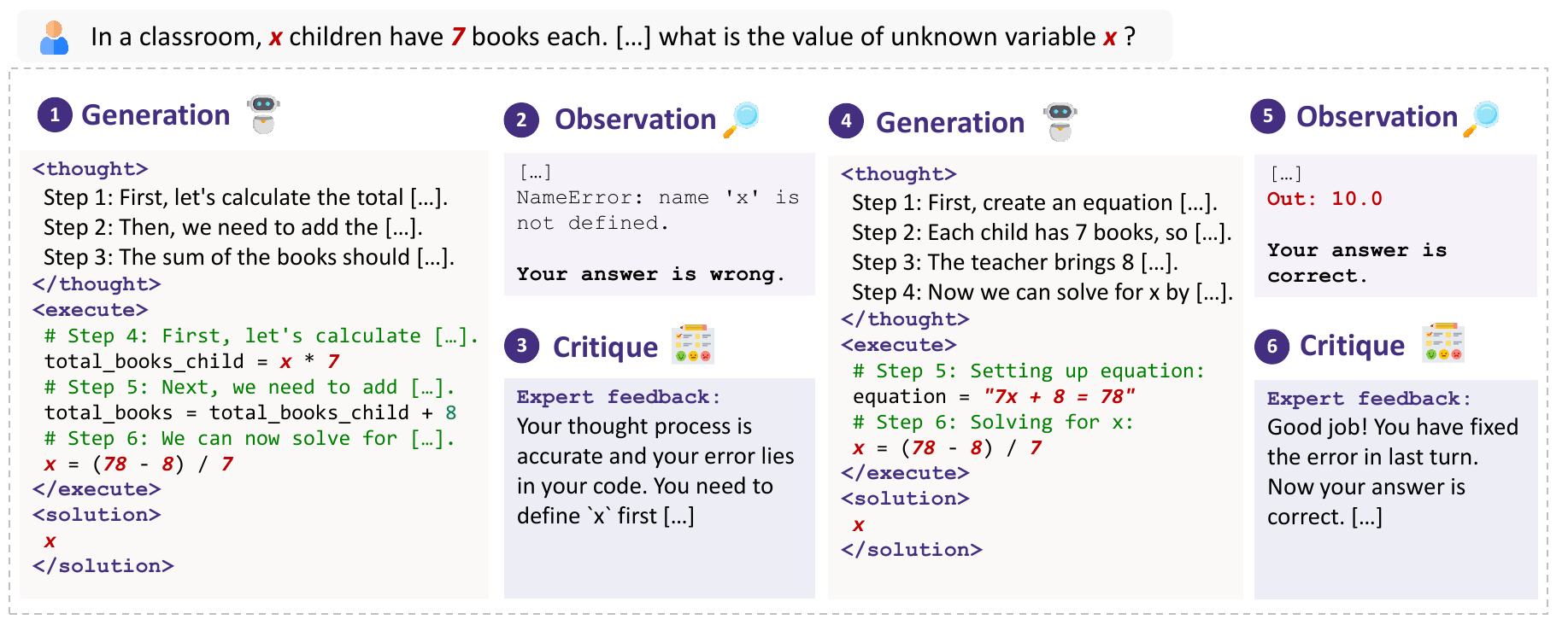}
    \vspace{-23pt}
    \caption{An illustrative example of an \dataname trajectory over two turns.
    In each turn, the actor model generates step-by-step reasoning chains, and the environment and the critique model provide observations and textual critique respectively.
    }
    \label{fig:ultraiteract}
    \vspace{-15pt}
\end{figure*}

\subsection{Decomposition and Interaction at Each Turn}
\label{sec:trajectory_generation}
\looseness=-1
Figure \ref{fig:ultraiteract} provides an illustrative example.
In what follows, we connect the actor model with a Python interpreter as the ``environment''.
Unless otherwise specified, we use GPT-3.5 Turbo as the actor model.

\looseness=-1
Following \citet{Wang2024ExecutableCA},
the actor model first decomposes the input problem into several sub-problems and then solves each by generating Python code pieces as actions and using the environment to execute them.
To promote solution diversity, the actor model randomly samples one reasoning schema in the form of either CoT \citep{Wei2022ChainOT} or modularization programming \citep{Qian2023CREATORTC, Yuan2023CRAFTCL}.
The actor then generates actions in text or code to solve each sub-problem, with each step being marked by explicit notations.

Multi-turn interactions with the environment are often necessary to solve challenging problems \citep{Wang2023MINTEL}. 
To improve such capabilities of the models, \dataname collects trajectories in which the actor model interacts with the environment and a critique model (a proxy for user) and refines its action based on their feedback.

\looseness=-1
The environment receives an action from the actor model along with the interaction history, and then the code interpreter returns two kinds of ``\textit{Observation}'': (1) Python execution results, either program outputs or error traceback messages; (2) binary feedback, indicating whether the solution is correct or not.
Then, the observations along with the history will be passed to a \textit{critique} model, which locates the errors and provides suggestions for improvements.
To avoid potential bias introduced by self-correction \citep{Wang2023MINTEL, Xu2024PerilsOS}, we adopt a stronger model, GPT-4, as the critique and ensure critique quality by providing GPT-4 with ground truth answers as references.

This procedure resembles \citet{Wang2024ExecutableCA}. 
However, we adopt more diverse reasoning patterns to teach LLMs to learn rationales rather than simply memorizing answers \citep{Mitra2023Orca2T}, and learn to create and use tools \citep{Qian2023CREATORTC, Yuan2023CRAFTCL, Qin2023ToolLLMFL}. Besides, we believe that it is important for LLMs to learn from the feedback provided by the critique rather than solely from observations of the environment.

\subsection{Preference Trees Facilitates Preference Learning Across Multiple Turns}
\label{sec:preference_data}

\looseness=-1
Unlike open-ended conversations, where human preference is ambiguous and challenging to specify, many reasoning tasks have clear and objective preferences for \textit{correct} actions. 
The preference annotation is threfore an evaluation of the correctness of the solutions \textit{conditioning ground truth ones}, which come with the datasets in \dataname.
This eliminates the need for human or LLM-based preference annotation and ensures high data quality.
To facilitate preference learning, \dataname pairs correct and incorrect actions.

\looseness=-1
\textbf{Sampling Paired Correct and Incorrect Actions at Each Turn.}
For each instruction in \dataname, we sample, from the actor model, a pair of correct and incorrect actions following \S\ref{sec:trajectory_generation}.
We follow \cite{Cui2023UltraFeedbackBL} to sample the pair from different actor models to ensure response diversity.
To prevent models from exploiting shortcuts based on surface features, we exclude instances that fail to pass the Python syntax check.

Certain challenging problems in \dataname pose difficulties in obtaining correct actions, even using strong actors such as GPT-4, with nearly zero \textit{pass@100} accuracies.
To improve the pass rates of the actor models while keeping the expense under control, we sequentially take the following steps.
(1) Directly sampling 20 actions and randomly keeping a correct one, if any. 
(2) If no correct action is obtained, we repeat the above process up to three times,
progressively switching from more cost-effective models to the strong yet expensive GPT-4 Turbo.
(3) For the remaining difficult problems where no correct action is acquired after the previous two steps, we provide the actor with ground-truth rationales and answers, and then apply various techniques to elicit correct actions. 
The specific information provided and the techniques applied vary depending on the tasks (Appendix \ref{sec:correct_action_generation_details}).

\looseness=-1
\textbf{Tree-structured Action Pairs Across Multiple Turns.} 
After each turn, the correct action concludes its trajectory.
We expand the \textit{incorrect} action into the next turn, and have the actor interact with the environment and the critique to refine its solution (\S\ref{sec:trajectory_generation}).
We then repeat the procedures introduced earlier in this section to collect an additional action pair.
By expanding the \textit{incorrect} action, \dataname can provide data to help models learn from feedback, and collect multiple action pairs for preference learning across multiple turns.

Conceptually, for every instruction, \dataname constructs a binary preference tree with each action being a node (Figure~\ref{fig:tree}).
We cap the tree at a maximum of five turns.

\textbf{Additional Instruction-action Pairs for Challenging Problems.}
We believe the challenging instructions that make it to step (3) above can provide valuable training signals.
Therefore, for a subset of these problems with multiple ground truth solutions, we further sample additional correct actions to cover all ground truths.
Accordingly, we further sample incorrect actions to pair with these additional correct actions, so that they can be used in both supervised fine-tuning and preference learning.

With the tree-structured data, \dataname enables comparisons at every turn, in contrast to comparing only at the last turn \citep{Bai2022TrainingAH}, and thus can improve the models' interaction ability.
Closing this section, Table \ref{tab:stats} summarizes some statistics of \dataname, and more details are in Appendix \ref{sec:stats_breakdown}.





\input{tabs/stats_new}

%% file: tabs/stats_new.tex
\begin{table*}[tbh]
\vspace{-15pt}
\caption{Some statistics of \dataname. }
\resizebox{\textwidth}{!}{

\begin{tabular}{@{}lcccccccccccc@{}}
\toprule
\multirow{2}{*}{\textbf{Task}}   & \multicolumn{2}{c}{\textbf{Type}}                         & \multirow{2}{*}{\textbf{\# Instructions}} & \multicolumn{5}{c}{\textbf{\# Turns per Traj.}}                     & \multirow{2}{*}{\textbf{\begin{tabular}[c]{@{}c@{}}\# Tokens \\ per Traj.\end{tabular}}} & \multirow{2}{*}{\textbf{\begin{tabular}[c]{@{}c@{}}Avg. \# Traj \\ per Ins.\end{tabular}}} & \multirow{2}{*}{\textbf{\begin{tabular}[c]{@{}c@{}}Total \\ \# Pairs\end{tabular}}} & \multirow{2}{*}{\textbf{\begin{tabular}[c]{@{}c@{}}\# Correct \\ Answers\end{tabular}}} \\ \cmidrule(lr){2-3} \cmidrule(lr){5-9}
                                 & \textbf{w/ Interaction?}    & \textbf{w/ Tool?}           &                                           & \textbf{T1} & \textbf{T2} & \textbf{T3} & \textbf{T4} & \textbf{T5} &                                                                                          &                                                                                            &                                                                                     &                                                                                         \\ \midrule
\multirow{4}{*}{\textbf{Math}}   & \Checkmark   & \Checkmark   & 22,928                                    & 10,440      & 4,122       & 1,898       & 904         & 5,564       & 1,750.0                                                                                  & 1.0                                                                                        & 42,780                                                                              & 68,033                                                                                  \\
                                 & \XSolidBrush   & \Checkmark & 2,757                                     & 16,154      & -           & -           & -           & -           & 439.1                                                                                    & 5.9                                                                                        & 13,217                                                                              & 16,154                                                                                  \\
                                 & \Checkmark &  \XSolidBrush  & 22,639                                    & 10,708      & 3,521       & 1,459       & 723         & 6,228       & 1,521.9                                                                                  & 1.0                                                                                        & 44,750                                                                              & 62,182                                                                                  \\
                                 & \XSolidBrush & \XSolidBrush & 2,083                                     & 16,348      & -           & -           & -           & -           & 538.1                                                                                    & 7.8                                                                                        & 12,624                                                                              & 16,348                                                                                  \\ \midrule
\multirow{2}{*}{\textbf{Coding}} & \Checkmark   & -                           & 20,463                                    & 13,265      & 2,584       & 987         & 379         & 3,248       & 1,728.5                                                                                  & 1.0                                                                                        & 18,106                                                                              & 22,215                                                                                  \\
                                 & \XSolidBrush & -                           & 8,495                                     & 92,618      & -           & -           & -           & -           & 1,070.4                                                                                  & 5.5                                                                                        & 78,634                                                                              & 92,618                                                                                  \\ \midrule
\multirow{2}{*}{\textbf{Logic}}  & \Checkmark   & \Checkmark   & 2,086                                     & 1,685       & 298         & 72          & 8           & 23          & 1,299.8                                                                                  & 1.0                                                                                        & 1,750                                                                               & 2,198                                                                                   \\
                                 & \Checkmark   & \XSolidBrush & 4,467                                     & 2,453       & 1,674       & 340         & 0           & 0           & 1,266.7                                                                                  & 1.0                                                                                        & 7,958                                                                               & 7,231                                                                                   \\ \midrule
\textbf{Total}                   & -                           & -                           & 85,918                                    & 163,671     & 12,199      & 4,756       & 2,014       & 15,063      & 1,201.8                                                                                  & 2.3                                                                                        & 219,819                                                                             & 286,979                                                                                 \\ \bottomrule
\end{tabular}
}
\label{tab:stats}

\end{table*}

%% file: sections/_model_intro.tex
\section{\modelname: State-of-the-art Open LLMs  in Reasoning}
\looseness=-1
\dataname helps us develop \modelname, a suite of LLMs and a reward model (RM).

\looseness=-1
\textbf{Supervised Fine-Tuning.}
\smallsft is fine-tuned from Mistral-7B~\citep{jiang2023mistral} and \bigsft from CodeLLaMA-70B~\citep{roziere2023code}. 
First, we perform SFT using all correct actions (287K) in \dataname. 
We find it yields better performance to discard interaction history and train only on correct leaf nodes in each tree. 
To improve general instruction-following ability, we include into our SFT data mixture UltraChat \citep{Ding2023EnhancingCL}, ShareGPT\footnote{\url{https://huggingface.co/datasets/openchat/openchat_sharegpt4_dataset}}, and OpenOrca \citep{OpenOrca}. Please find mixture ratios in Appendix \ref{sec:training_details}.

\textbf{Perference Learning.}
Based on \sft models, we explore three preference learning algorithms, DPO~\citep{Rafailov2023DirectPO}, KTO~\citep{Ethayarajh2024KTOMA}, and NCA~\citep{Chen2024NoiseCA}.
Differently from SFT, here we include all multi-turn trajectory pairs in our \dataname (220K) and include all UltraFeedback~\citep{Cui2023UltraFeedbackBL} pairs (340K).

\looseness=-1
\textbf{Reward Modeling.}
Similarly to the preference learning, we use all 220K multi-turn trajectory pairs from \dataname;
it is further augmented with the 240K single-turn action pairs from \dataname.
More details are in the Appendix \ref{sec:training_details}.
We include all 340K pairs from UltraFeedback and one pair for each instruction from UltraSafety \citep{Guo2024ControllablePO}, totaling 3K.
\rewardmodel is initialized from \smallsft with a new linear layer.

Our findings in \S\ref{sec:analysis} indicate that the absolute values of rewards make a big difference in the models' reasoning performance.
We therefore augment the established Bradley-Terry (BT) objective $\mathcal{L}_{\text{BT}}$ with an additional term $\mathcal{L}_{\text{DR}}$ to \textbf{d}i\textbf{r}ectly increase the reward of the chosen actions for instances from \dataname, and decrease those of the rejected ones:


\vspace{-20pt}
\begin{equation*}
\mathcal{L}_{\text{\tiny\dataname}} = 
\underbrace{
-\log\Bigl(\sigma \bigl(r_{\theta}(x, y_{c})-r_{\theta}(x, y_{r})\bigr)\Bigr)
}_\text{$\mathcal{L_{\text{BT}}}$: optimize relative rewards} 
\underbrace{
-\log\Bigl(\sigma \bigl(r_{\theta}(x, y_{c})\bigr)\Bigr) -\log\Bigl(\sigma \bigl(-r_{\theta}(x, y_{r})\bigr)\Bigr)
}_\text{$\mathcal{L_{\text{DR}}}$: increase $r_{\theta}(x, y_{c})$ and decrease $r_{\theta}(x, y_{r})$}
\end{equation*}

%

\vspace{-10pt}
For instances from other datasets, we train with $\mathcal{L_{\text{BT}}}$.
$\theta$ denotes the reward model's parameters, $r_{\theta}\left(\cdot\right)$ and $r_{\theta}\left(x, y_{r}\right)$ the rewards on the chosen and rejected actions respectively.
Our ablation study demonstrates the importance of both $\mathcal{L}_{\text{BT}}$ and 
$\mathcal{L}_{\text{DR}}$.


%% file: sections/4_experiment.tex
\section{Evaluation of \smallmodel and \bigmodel}
\label{sec:exp}

\looseness=-1
\textbf{Evaluation Setup.}
We consider both single-turn and multi-turn reasoning. 
For single-turn evaluation, we consider HumanEval \citep{chen2021evaluating}, MBPP \citep{austin2021program}, and LeetCode \citep{deepseek-coder} for coding, GSM-Plus \citep{Li2024GSMPlusAC}, MATH, TheoremQA \citep{Chen2023TheoremQAAT}, SVAMP \citep{Patel2021SVAMP}, and ASDiv \citep{Miao2020ASDiv} for math, and BBH-Hard \citep{suzgun2022challenging} for reasoning. 
We evaluate with pass@1 accuracy. 
We also use IFEval \citep{Zhou2023InstructionFollowingEF} to assess the instruction-following ability and report the prompt-level loose score.
For multi-turn evaluation, we adopt MINT \citep{Wang2023MINTEL} and only consider the coding and math problems. We report the success rate at Turn 5.
Please find further details on evaluation setups and evaluations beyond reasoning in Appendix \ref{sec:additional_lm_eval}.


As shown in Table \ref{tab:model_list}, we compare our \modelname  with general-purpose models, and those specialized in coding and math of various sizes.
\input{tabs/model_list}
We also summarize the results of GPT-3.5 Turbo and GPT-4 reported in previous works.

\input{tabs/sft_new}
\subsection{Results}
Results are shown in Table \ref{tab:sft}. We summarize the takeaways as follows:

\looseness=-1
\textbf{\modelname, both the 7B and 70B variants, achieve the best overall performance among open-source models of similar sizes.
\modelname even outperform specialized models in corresponding domains in many cases.}
Notably, \smallmodel outperforms baselines that are 5$\times$ larger and \textbf{\bigmodel achieves better performance than GPT-3.5 Turbo}. 
\modelname's instruction-following performance is among the best general-purpose models, substantially better than specialized ones.


\textbf{Preference learning with \dataname can further improve the performance, especially in math and the multi-turn ability.}
KTO and NCA consistently improve the models' performance in all five math benchmarks and mult-turn evaluations, while their effects vary in others. 
Since SFT  models only use the single-turn data from \dataname while preference learning uses the multi-turn ones, the improvements in interaction ability should also be attributed to \dataname rather than the algorithms alone.
Surprisingly, we observe that \textbf{DPO hurts model performance on most benchmarks}.
DPO training of our 70B model fails since the rewards go down to $-\infty$. 
We analyze this phenomenon in \S \ref{sec:reward_pattern}.




\section{Evaluation of \rewardmodel}
\label{sec:rm}

\looseness=-1
\textbf{Evaluation Setup.}
We evaluate \rewardmodel on three RM benchmarks, RewardBench~\citep{Lambert2024RewardBenchER}, AutoJ \citep{Li2023GenerativeJF}, and MT-Bench~\citep{Zheng2023JudgingLW}.
Aiming for a more realistic OOD evalation, we exclude the ``prior sets'' split from RewardBench, since many baselines train on the datasets that this split contains.
We compare with PairRM \citep{Jiang2023LLMBlenderEL}, Starling-RM-7B/34B \citep{starling2023}, UltraRM-13B \citep{Cui2023UltraFeedbackBL}, GPT-3.5 Turbo, and GPT-4.
To further explore \rewardmodel's potential in improving models' performance through reranking, we use it to rerank Mistral-7B-Instruct-v0.2's responses on HumanEval, MBPP, GSM8K, and MATH. 
We report the results of random sampling, self-consistency, and Starling-RM-34B as baselines.

\subsection{Results}

Table \ref{tab:rm_benchmarks} summarizes reward modeling performance, and Figure \ref{fig:bon} plots some reranking results with others in Appendix \ref{sec:bon}.

\textbf{\rewardmodel stands out as the best 7B RM overall, and achieves similar or better performance than much larger baselines.
Particularly, it outperforms GPT-4 in certain tasks. } \rewardmodel achieves a better correlation with human experts than all existing models on AutoJ and MT-Bench, and it achieves comparable performance to the 5$\times$ larger Starling-RM-34B on RewardBench. 
On RewardBench, \rewardmodel outperforms all baselines on the ``Chat-Hard'' split while achieving very competitive performance on the ``Reasoning'' split. 
Across the AutoJ splits, \rewardmodel outperforms nearly all existing models, with the only exception being GPT-4's results on Coding.

\textbf{Our training objective is beneficial in improving RM performance on hard problems and reasoning.} Table \ref{tab:rm_benchmarks} shows that optimizing $\mathcal{L}_{\text{DR}}$ improves RM's reasoning ability, but BT modeling is still beneficial in equipping RM with abilities in general chatting as suggested in the ``Chat-Hard'' column, though its effect on reasoning may vary.

\textbf{\dataname is compatible with other datasets like UltraFeedback and UltraSafety, and mixing these datasets can balance different RM abilities.} 
Improving RM's capabilities in reasoning with \dataname does not sacrifice others, which indicates that \dataname can be a great ingredient for the training data mixture of reward models.

\begin{figure}[tbh]
    \vspace{-5pt}
    \centering
    \includegraphics[width=0.24\textwidth]{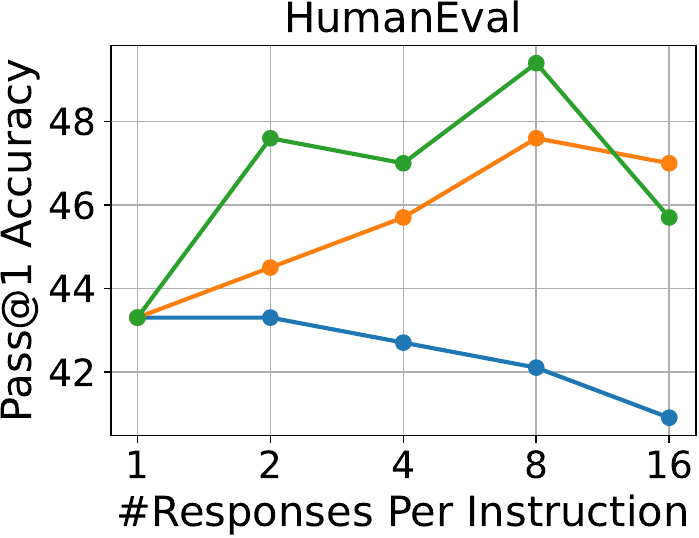}
    \hfill
    \includegraphics[width=0.24\textwidth]{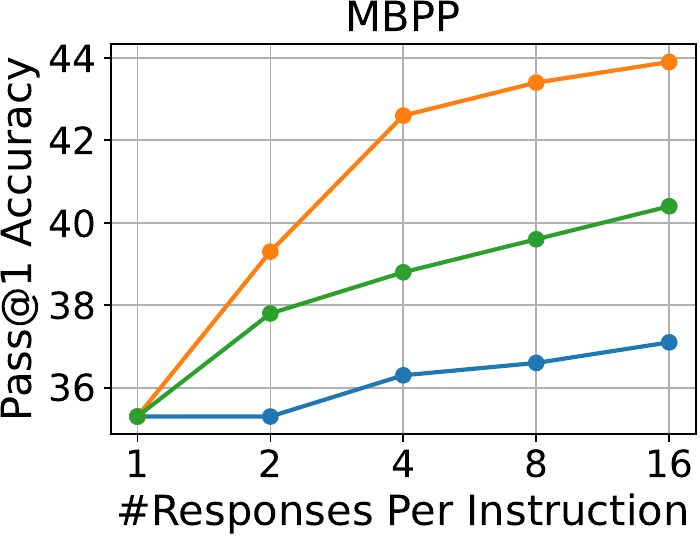}
    \hfill
    \includegraphics[width=0.24\textwidth]{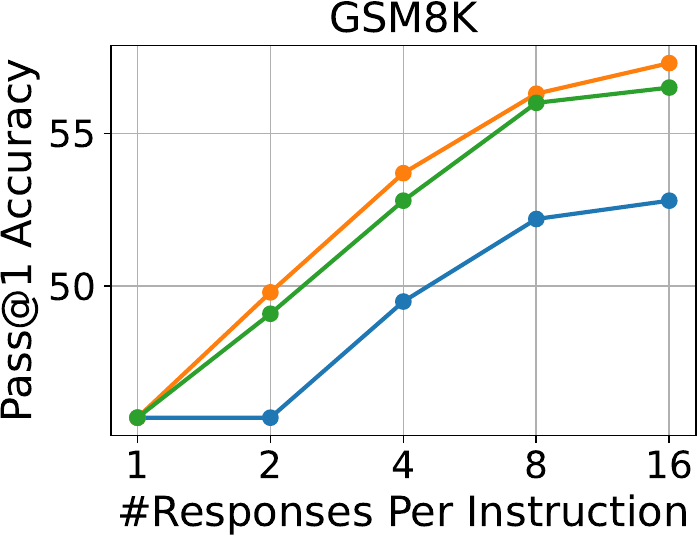}
    \hfill
    \includegraphics[width=0.25\textwidth]{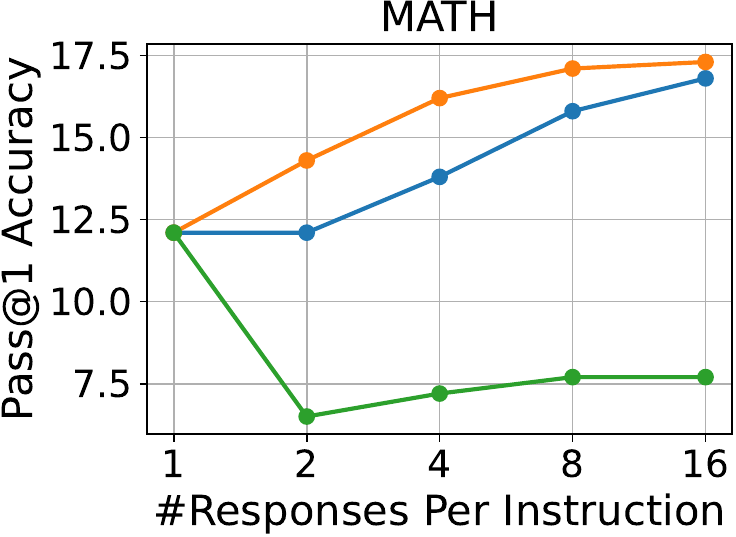}
    \\
    \centering
    \includegraphics[width=0.5\textwidth]{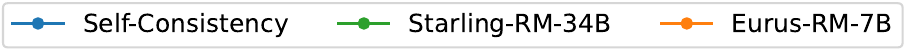}
    \vspace{-12pt}
    \caption{Results on reranking Mistral-7B-Instruct-v0.2's responses. Full results in Table \ref{tab:best_of_n}.}
    \label{fig:bon}
    \vspace{-10pt}
\end{figure}

\textbf{\rewardmodel improves LLMs' reasoning performance by a large margin through reranking.} 
\rewardmodel consistently improves pass@1 accuracy across all tasks and performs better than 5$\times$ larger baseline Starling-RM-34B. Also, \rewardmodel's reranking performance scales well with \#responses per instruction, except a slight decrease in HumanEval when increasing response number form 8 to 16.
In contrast, Starling-RM-34B suffers from severe performance drop on HumanEval and it consistently hurts model accuracy on MATH.

\input{tabs/rm_benchmarks}

\input{sections/5_analysis}

%% file: tabs/model_list.tex
\begin{table}[tb]
\caption{Open-source LLM baselines that we compare to.}
\label{tab:model_list}
\small
\resizebox{\textwidth}{!}{

\begin{tabular}{@{}l p{14.5cm}@{}}
\toprule

\bf Type& \multicolumn{1}{c}{\textbf{Models}} \\ \midrule
\multirow{3}{*}{\bf General Purpose}     & Mistral-7B-Instruct-v0.2~\citep{jiang2023mistral}, Zephyr-7B-$\beta$~\citep{tunstall2023zephyr}, OpenChat-3.5-1210~\citep{wang2023openchat}, Starling-LM-7B-$\alpha$~\citep{starling2023}, Mixtral-8x7B-Instruct~\citep{jiang2023mistral}, DeepSeek-LLM-67B-Chat~\citep{deepseek-llm}, QWen1.5-72B-Chat~\citep{qwen} \\ \midrule
\multirow{2}{*}{\bf Coding} & Magicoder-S-DS-6.7B~\citep{magicoder}, OpenCodeInterpreter (OpenCI for short, DS-6.7B/CL-70B)~\citep{zheng2024opencodeinterpreter}, DeepSeek-Coder-33B-Instruct~\citep{deepseek-coder}, and CodeLLaMA-70B-Instruct\citep{roziere2023code}.                         \\ \midrule
\multirow{2}{*}{\bf Math}   & MAmmoTH-7B-Mistral~\citep{yue2023mammoth}, WizardMath-7B-v1.1~\citep{luo2023wizardmath}, OpenMath (Mistral-7B/CodeLLaMA-70B)~\citep{toshniwal2024openmath}.                        \\ \bottomrule
\end{tabular}
}
\vspace{-30pt}
\end{table}

%% file: tabs/sft_new.tex
\begin{table}[tbh]
\caption{Overall performance. All test sets except MATH are out-of-distribution to our models and most baselines. 
MAmmoTH, OpenChat, and Starling-LM have been trained on TheoremQA test sets. 
We \sout{strikethrough} the contaminated numbers. }
\label{tab:sft}
\resizebox{\textwidth}{!}{
\begin{tabular}{@{}lccccccccccccc@{}}
\toprule
\multicolumn{1}{l|}{}                                               & \multicolumn{3}{c|}{\textbf{Coding}}                                                       & \multicolumn{5}{c|}{\textbf{Math}}                                                                                         & \multicolumn{1}{c|}{\textbf{Reasoning}}                    & \multicolumn{1}{c|}{\textbf{Ins-Following}}       & \multicolumn{2}{c|}{\textbf{Multi-Turn}}                                   &                                 \\ \cmidrule(lr){2-13}
\multicolumn{1}{l|}{\multirow{-2}{*}{\textbf{Model}}}               & HumanE.     & MBPP          & \multicolumn{1}{c|}{LeetC.}                                  & GSM-Plus      & MATH          & Theo.QA     & SVAMP         & \multicolumn{1}{c|}{ASDiv}                                 & \multicolumn{1}{c|}{BBH}                                   & \multicolumn{1}{c|}{IFEval}                       & Code          & \multicolumn{1}{c|}{Math}                                  & \multirow{-2}{*}{\textbf{Avg.}} \\ \midrule
\multicolumn{14}{c}{$\sim$7B}                                                                                                                                                                                                                                                                                                                                                                                                                                                                                                       \\ \midrule
\multicolumn{1}{l|}{\textbf{Mistral-7B-Instruct-v0.2}}                 & 39.0          & 30.8          & \multicolumn{1}{c|}{6.1}                                   & 15.7          & 9.5           & 8.5           & 42.9          & \multicolumn{1}{c|}{49.5}                                  & \multicolumn{1}{c|}{62.4}                                  & \multicolumn{1}{c|}{44.4}                         & 7.4           & \multicolumn{1}{c|}{26.2}                                  & 28.5                            \\
\multicolumn{1}{l|}{\textbf{Zephyr-7B-$\beta$}}                                & 29.3          & 35.8          & \multicolumn{1}{c|}{2.2}                                   & 23.3          & 5.0           & 7.8           & 19.1          & \multicolumn{1}{c|}{28.0}                                  & \multicolumn{1}{c|}{61.8}                                  & \multicolumn{1}{c|}{39.7}                         & 5.2           & \multicolumn{1}{c|}{16.9}                                  & 22.8                            \\
\multicolumn{1}{l|}{\textbf{OpenChat-3.5-1210}}                     & 64.0          & 61.7          & \multicolumn{1}{c|}{11.7}                                  & 46.7          & 28.1          & \sout{19.1}          & 75.4          & \multicolumn{1}{c|}{77.0}                                  & \multicolumn{1}{c|}{67.0}                                  & \multicolumn{1}{c|}{\textbf{50.3}}                & 21.3          & \multicolumn{1}{c|}{32.4}                                  & \sout{46.2}                            \\
\multicolumn{1}{l|}{\textbf{Starling-LM-7B-$\alpha$}}                           & 46.3          & 51.1          & \multicolumn{1}{c|}{8.9}                                   & 23.7          & 21.5          & \sout{12.0}          & 26.3          & \multicolumn{1}{c|}{39.8}                                  & \multicolumn{1}{c|}{67.1}                                  & \multicolumn{1}{c|}{26.1}                         & 18.4          & \multicolumn{1}{c|}{28.9}                                  & \sout{30.8}                            \\
\multicolumn{1}{l|}{\textbf{Magicoder-S-DS-6.7B}}                        & 75.6          & \textbf{70.4} & \multicolumn{1}{c|}{\textbf{23.9}}                         & 16.4          & 19.9          & 13.1          & 61.6          & \multicolumn{1}{c|}{62.8}                                  & \multicolumn{1}{c|}{57.0}                                  & \multicolumn{1}{c|}{21.1}                         & \textbf{27.9} & \multicolumn{1}{c|}{8.0}                                   & 38.1                            \\
\multicolumn{1}{l|}{\textbf{OpenCI-DS-6.7B}}                        & \textbf{76.8} & 66.2          & \multicolumn{1}{c|}{16.1}                                  & 41.5          & 31.6          & 16.1          & 74.5          & \multicolumn{1}{c|}{79.8}                                  & \multicolumn{1}{c|}{53.9}                                  & \multicolumn{1}{c|}{22.6}                         & 5.9           & \multicolumn{1}{c|}{1.3}                                   & 40.5                            \\
\multicolumn{1}{l|}{\textbf{MAmmoTH-7B-Mistral}}                       & 24.4          & 42.4          & \multicolumn{1}{c|}{7.2}                                   & 40.1          & 36.0          & \sout{26.3}          & 60.7          & \multicolumn{1}{c|}{72.3}                                  & \multicolumn{1}{c|}{57.7}                                  & \multicolumn{1}{c|}{34.9}                         & 3.7           & \multicolumn{1}{c|}{6.7}                                  & \sout{34.4}                            \\
\multicolumn{1}{l|}{\textbf{WizardMath-7B-v1.1}}                       & 50.0          & 53.9          & \multicolumn{1}{c|}{6.7}                                   & 54.6          & 30.0          & 16.5          & 57.8          & \multicolumn{1}{c|}{73.5}                                  & \multicolumn{1}{c|}{64.4}                                  & \multicolumn{1}{c|}{22.6}                         & 16.2          & \multicolumn{1}{c|}{8.9}                                  & 37.9                            \\
\multicolumn{1}{l|}{\textbf{OpenMath-Mistral-7B}}                      & 33.5          & 46.6          & \multicolumn{1}{c|}{11.7}                                  & \textbf{59.4} & \textbf{39.1} & 13.1          & 83.4          & \multicolumn{1}{c|}{79.8}                                  & \multicolumn{1}{c|}{58.6}                                  & \multicolumn{1}{c|}{15.0}                         & 2.9           & \multicolumn{1}{c|}{5.3}                                   & 37.4                            \\
\rowcolor[HTML]{DAE8FC} 
\multicolumn{1}{l|}{\cellcolor[HTML]{DAE8FC}\textbf{\smallsft}}  & 55.5          & 59.1          & \multicolumn{1}{c|}{\cellcolor[HTML]{DAE8FC}20.0}          & 52.1          & 32.6          & 20.0          & 82.2          & \multicolumn{1}{c|}{\cellcolor[HTML]{DAE8FC}84.1}          & \multicolumn{1}{c|}{\cellcolor[HTML]{DAE8FC}64.6}          & \multicolumn{1}{c|}{\cellcolor[HTML]{DAE8FC}44.0} & 15.4          & \multicolumn{1}{c|}{\cellcolor[HTML]{DAE8FC}28.4}          & 46.5                            \\
\rowcolor[HTML]{DAE8FC} 
\multicolumn{1}{l|}{\cellcolor[HTML]{DAE8FC}\quad\textbf{+ DPO}}         & 50.6          & 52.1          & \multicolumn{1}{c|}{\cellcolor[HTML]{DAE8FC}8.3}           & 51.0          & 28.3          & \textbf{20.9} & 78.7          & \multicolumn{1}{c|}{\cellcolor[HTML]{DAE8FC}83.8}          & \multicolumn{1}{c|}{\cellcolor[HTML]{DAE8FC}65.0}          & \multicolumn{1}{c|}{\cellcolor[HTML]{DAE8FC}42.5} & 20.6          & \multicolumn{1}{c|}{\cellcolor[HTML]{DAE8FC}32.4}          & 44.5                           \\
\rowcolor[HTML]{DAE8FC} 
\multicolumn{1}{l|}{\cellcolor[HTML]{DAE8FC}\quad\textbf{+ KTO}}         & 56.1          & 58.6          & \multicolumn{1}{c|}{\cellcolor[HTML]{DAE8FC}18.9}          & 55.0          & 33.2          & 20.6          & 84.4          & \multicolumn{1}{c|}{\cellcolor[HTML]{DAE8FC}85.0}          & \multicolumn{1}{c|}{\cellcolor[HTML]{DAE8FC}\textbf{67.6}} & \multicolumn{1}{c|}{\cellcolor[HTML]{DAE8FC}43.1} & 19.1          & \multicolumn{1}{c|}{\cellcolor[HTML]{DAE8FC}\textbf{43.6}} & \textbf{48.8}                   \\
\rowcolor[HTML]{DAE8FC} 
\multicolumn{1}{l|}{\cellcolor[HTML]{DAE8FC}\quad\textbf{+ NCA}}         & 55.5          & 60.2          & \multicolumn{1}{c|}{\cellcolor[HTML]{DAE8FC}14.4}          & 54.9          & 34.2          & \textbf{20.9}          & \textbf{84.6} & \multicolumn{1}{c|}{\cellcolor[HTML]{DAE8FC}\textbf{85.4}} & \multicolumn{1}{c|}{\cellcolor[HTML]{DAE8FC}64.3}          & \multicolumn{1}{c|}{\cellcolor[HTML]{DAE8FC}42.7} & 21.3          & \multicolumn{1}{c|}{\cellcolor[HTML]{DAE8FC}38.7}          & 48.1                            \\ \midrule
\multicolumn{14}{c}{$\sim$40B}                                                                                                                                                                                                                                                                                                                                                                                                                                                                                                      \\ \midrule
\multicolumn{1}{l|}{\textbf{Mixtral-8x7B-Instruct}}                 & 50.6          & 50.1          & \multicolumn{1}{c|}{5.6}                                   & \textbf{49.6} & \textbf{25.9} & 20.4          & 66.4          & \multicolumn{1}{c|}{68.8}                                  & \multicolumn{1}{c|}{\textbf{73.5}}                         & \multicolumn{1}{c|}{\textbf{48.8}}                & 12.5          & \multicolumn{1}{c|}{\textbf{37.3}}                         & 42.5                            \\
\multicolumn{1}{l|}{\textbf{DeepSeek-Coder-33B-Ins}}                    & \textbf{82.3} & \textbf{73.9} & \multicolumn{1}{c|}{\textbf{27.8}}                         & 29.5          & 20.2          & \textbf{21.9} & \textbf{75.2} & \multicolumn{1}{c|}{\textbf{85.0}}                         & \multicolumn{1}{c|}{61.5}                                  & \multicolumn{1}{c|}{26.1}                         & \textbf{35.3} & \multicolumn{1}{c|}{21.8}                                  & \textbf{46.7}                   \\ \midrule
\multicolumn{14}{c}{$\sim$70B}                                                                                                                                                                                                                                                                                                                                                                                                                                                                                                      \\ \midrule
\multicolumn{1}{l|}{\textbf{CodeLLaMA-70B-Instruct}}                & 56.7          & 58.6          & \multicolumn{1}{c|}{14.4}                                  & 34.9          & 12.0          & 8.4           & 63.5          & \multicolumn{1}{c|}{70.1}                                  & \multicolumn{1}{c|}{74.5}                                  & \multicolumn{1}{c|}{24.0}                         & 3.7           & \multicolumn{1}{c|}{14.2}                                  & 36.3                            \\
\multicolumn{1}{l|}{\textbf{DeepSeek-LM-67B-Chat}}                  & 70.7          & 65.7          & \multicolumn{1}{c|}{20.0}                                  & 65.0          & 41.0          & 17.9          & 74.0          & \multicolumn{1}{c|}{84.0}                                  & \multicolumn{1}{c|}{78.9}                                  & \multicolumn{1}{c|}{52.7}                         & 30.9          & \multicolumn{1}{c|}{41.8}                                  & 53.5                            \\
\multicolumn{1}{l|}{\textbf{QWen1.5-72B-Chat}}                          & 71.3          & 56.9          & \multicolumn{1}{c|}{15.6}                                  & \textbf{65.4} & 43.4          & 18.5          & 79.5          & \multicolumn{1}{c|}{79.1}                                  & \multicolumn{1}{c|}{78.0}                                  & \multicolumn{1}{c|}{\textbf{53.4}}                & 27.2          & \multicolumn{1}{c|}{38.2}                                  & 52.2                            \\
\multicolumn{1}{l|}{\textbf{OpenCI-CL-70B}}                         & 77.4          & 71.7          & \multicolumn{1}{c|}{20.0}                                  & 46.1          & 29.2          & 18.8          & 76.1          & \multicolumn{1}{c|}{79.4}                                  & \multicolumn{1}{c|}{66.7}                                  & \multicolumn{1}{c|}{26.8}                         & 30.9          & \multicolumn{1}{c|}{12.0}                                  & 46.3                            \\
\multicolumn{1}{l|}{\textbf{OpenMath-CL-70B}}                          & 39.0          & 52.6          & \multicolumn{1}{c|}{15.0}                                  & 62.2          & \textbf{45.9} & 15.9          & 86.6          & \multicolumn{1}{c|}{82.8}                                  & \multicolumn{1}{c|}{59.9}                                  & \multicolumn{1}{c|}{15.7}                         & 14.0          & \multicolumn{1}{c|}{0.4}                                   & 40.8                            \\
\rowcolor[HTML]{DAE8FC} 
\multicolumn{1}{l|}{\cellcolor[HTML]{DAE8FC}\textbf{\bigsft}} & 75.6          & \textbf{74.2} & \multicolumn{1}{c|}{\cellcolor[HTML]{DAE8FC}\textbf{33.3}} & 58.1          & 40.6          & 28.0          & 86.3          & \multicolumn{1}{c|}{\cellcolor[HTML]{DAE8FC}88.5}          & \multicolumn{1}{c|}{\cellcolor[HTML]{DAE8FC}79.9}          & \multicolumn{1}{c|}{\cellcolor[HTML]{DAE8FC}49.2} & 31.6          & \multicolumn{1}{c|}{\cellcolor[HTML]{DAE8FC}40.4}          & 57.1                            \\
\rowcolor[HTML]{DAE8FC} 
\multicolumn{1}{l|}{\cellcolor[HTML]{DAE8FC}\quad\textbf{+ KTO}}         & 76.8          & 68.2          & \multicolumn{1}{c|}{\cellcolor[HTML]{DAE8FC}26.1}          & 62.2          & 41.3          & 30.6          & \textbf{90.4} & \multicolumn{1}{c|}{\cellcolor[HTML]{DAE8FC}89.0}          & \multicolumn{1}{c|}{\cellcolor[HTML]{DAE8FC}\textbf{80.8}} & \multicolumn{1}{c|}{\cellcolor[HTML]{DAE8FC}46.4} & \textbf{39.0} & \multicolumn{1}{c|}{\cellcolor[HTML]{DAE8FC}\textbf{49.8}} & 58.4                            \\
\rowcolor[HTML]{DAE8FC} 
\multicolumn{1}{l|}{\cellcolor[HTML]{DAE8FC}\quad\textbf{+ NCA}}         & \textbf{79.3} & 71.9          & \multicolumn{1}{c|}{\cellcolor[HTML]{DAE8FC}\textbf{33.3}} & 62.8          & 41.7          & \textbf{32.6} & 89.5          & \multicolumn{1}{c|}{\cellcolor[HTML]{DAE8FC}\textbf{90.3}} & \multicolumn{1}{c|}{\cellcolor[HTML]{DAE8FC}80.0}          & \multicolumn{1}{c|}{\cellcolor[HTML]{DAE8FC}49.2} & 38.2          & \multicolumn{1}{c|}{\cellcolor[HTML]{DAE8FC}39.6}          & \textbf{59.0}                   \\ \midrule
\multicolumn{14}{c}{Proprietary Models}                                                                                                                                                                                                                                                                                                                                                                                                                                                                                       \\ \midrule
\multicolumn{1}{l|}{\textbf{GPT-3.5 Turbo}}                         & 76.8          & 82.5          & \multicolumn{1}{c|}{23.3}                                  & 61.2          & 37.8          & 35.6          & 83.0          & \multicolumn{1}{c|}{90.6}                                  & \multicolumn{1}{c|}{70.1}                                  & \multicolumn{1}{c|}{56.6}                         & 29.4          & \multicolumn{1}{c|}{36.9}                                  & 57.0                            \\
\multicolumn{1}{l|}{\textbf{GPT-4}}                                 & \textbf{85.4} & \textbf{83.5} & \multicolumn{1}{c|}{\textbf{41.8}}                         & \textbf{85.6} & \textbf{69.7} & \textbf{52.4} & \textbf{94.8} & \multicolumn{1}{c|}{\textbf{92.6}}                         & \multicolumn{1}{c|}{\textbf{86.7}}                         & \multicolumn{1}{c|}{\textbf{79.7}}                & \textbf{59.6} & \multicolumn{1}{c|}{\textbf{65.8}}                         & \textbf{74.8}                   \\ \bottomrule
\end{tabular}
}
\end{table}

%% file: tabs/rm_benchmarks.tex
\begin{table}[]
\caption{Results on reward modeling benchmarks. UF: UltraFeedback; US: UltraSafety. The best performance in each benchmark is in bold and the second best one is underlined. Most baseline results are from \cite{Jiang2023LLMBlenderEL} and \cite{Lambert2024RewardBenchER}.}
\label{tab:rm_benchmarks}
\resizebox{\textwidth}{!}{
\begin{tabular}{@{}lccccclcccclc@{}}
\toprule
\multirow{2}{*}{~ Model}                       & \multicolumn{5}{c}{Reward Bench}                                 &  & \multicolumn{4}{c}{AutoJ}            &  & \multirow{2}{*}{MT-Bench} \\ \cmidrule(lr){2-6} \cmidrule(lr){8-11}
                                             & Chat & Chat-Hard & Safety & Reasoning & \multicolumn{1}{l}{Avg.} &  & Code & Math & Others & Overall       &  &                           \\ \midrule
\multicolumn{1}{l}{PairRM}           & 90.2 & 53.0      & 31.5   & 60.0      & 58.7                     &  & 58.3 & 52.8 & 58.9   & 59.1          &  & 59.0                      \\
\multicolumn{1}{l}{Starling-RM-7B}    & 98.0 & 43.4      & 88.6   & 74.6      & 76.2                     &  & 59.2 & 47.2 & 61.4   & 60.8          &  & 56.8                      \\
\multicolumn{1}{l}{Starling-RM-34B}         & 96.9 & 59.0      & 89.9   & 90.3      & \textbf{84.0}            &  & 65.8 & 54.2 & 62.3   & 62.6          &  & 60.4                      \\
\multicolumn{1}{l}{UltraRM-13B}       & 96.1 & 55.3      & 45.8   & 82.0      & 69.8                     &  & 55.0 & 43.1 & 59.6   & 59.9          &  & 56.0                      \\
\multicolumn{1}{l}{GPT-3.5 Turbo}      & -    & -         & -      & -         & -                        &  & 36.6 & 40.3 & 41.2   & 42.7          &  & 57.1                      \\
\multicolumn{1}{l}{GPT -4}                  & -    & -         & -      & -         & -                        &  & 69.2 & 51.4 & 61.4   & 61.9          &  & 63.9                      \\ \midrule
\multicolumn{1}{l}{\rewardmodel}      & 96.5 & 65.3      & 80.7   & 87.0      & \underline{82.4}               &  & 67.5 & 62.5 & 63.6   & 64.5          &  & \textbf{72.9}             \\
\multicolumn{1}{l}{w/o $\mathcal{L}_{\text{DR}}$}      & 96.4 & 59.9      & 79.5   & 77.5      & 78.3                     &  & 64.2 & 59.7 & 64.7   & 65.0    &  & \underline{72.8}                \\
\multicolumn{1}{l}{w/o $\mathcal{L}_{\text{BT}}$} & 96.8 & 58.5      & 83.8   & 84.2      & 80.8                     &  & 67.5 & 66.7 & 64.8   & \underline{65.6}          &  & 72.6                      \\
\multicolumn{1}{l}{w/o US}                  & 96.5 & 66.2      & 67.7   & 81.7      & 73.3                     &  & 66.7 & 61.1 & 65.0   & \textbf{65.7} &  & 72.6                      \\
\multicolumn{1}{l}{w/o UF + US}             & 95.1 & 61.1      & 63.7   & 73.4      & 78.0                     &  & 55.8 & 58.3 & 59.0   & 58.7          &  & 67.2                      \\ \bottomrule
\end{tabular}
}
\vspace{-12pt}
\end{table}

%% file: sections/5_analysis.tex
\section{Analysis}
\label{sec:analysis}

\begin{figure}[tbh]
    \vspace{-8pt}
    \centering
    \includegraphics[width=0.32\textwidth]{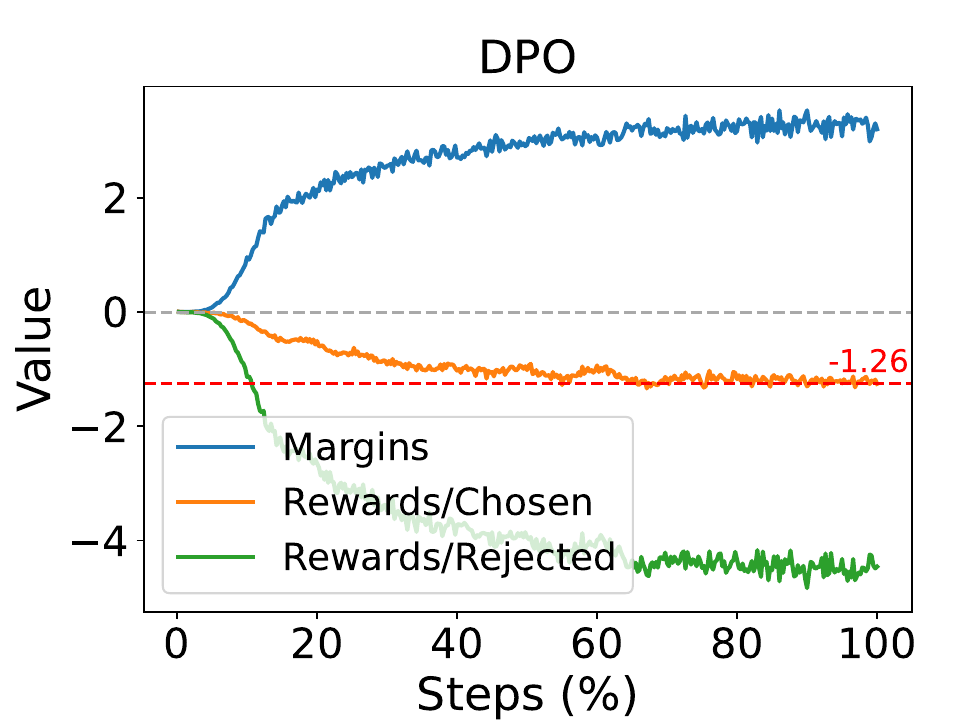}
    \hfill
    \includegraphics[width=0.32\textwidth]{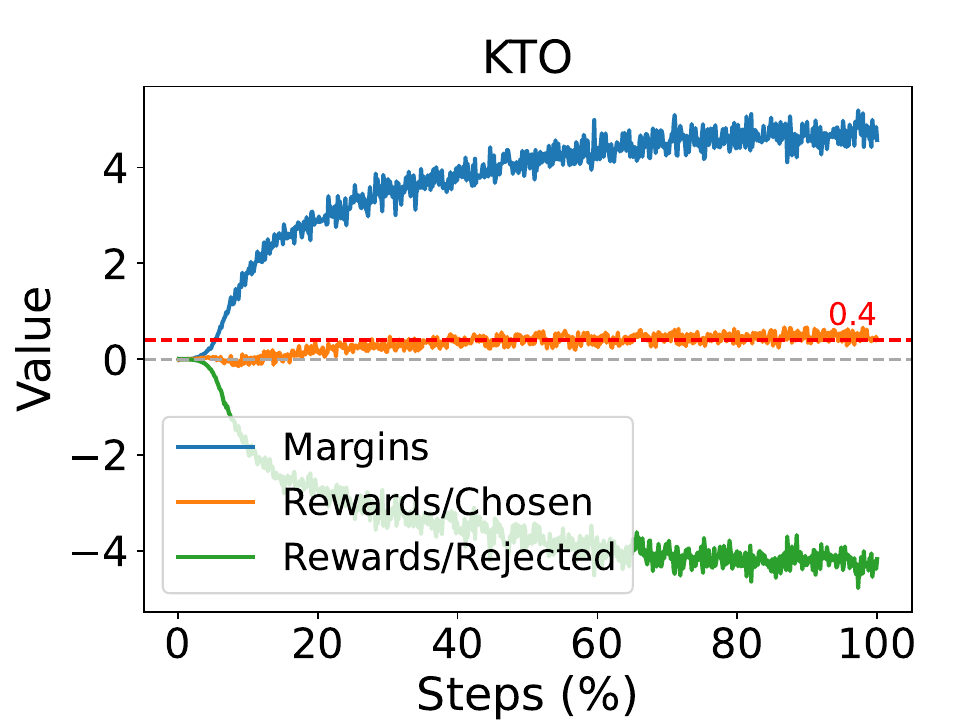}
    \hfill
    \includegraphics[width=0.32\textwidth]{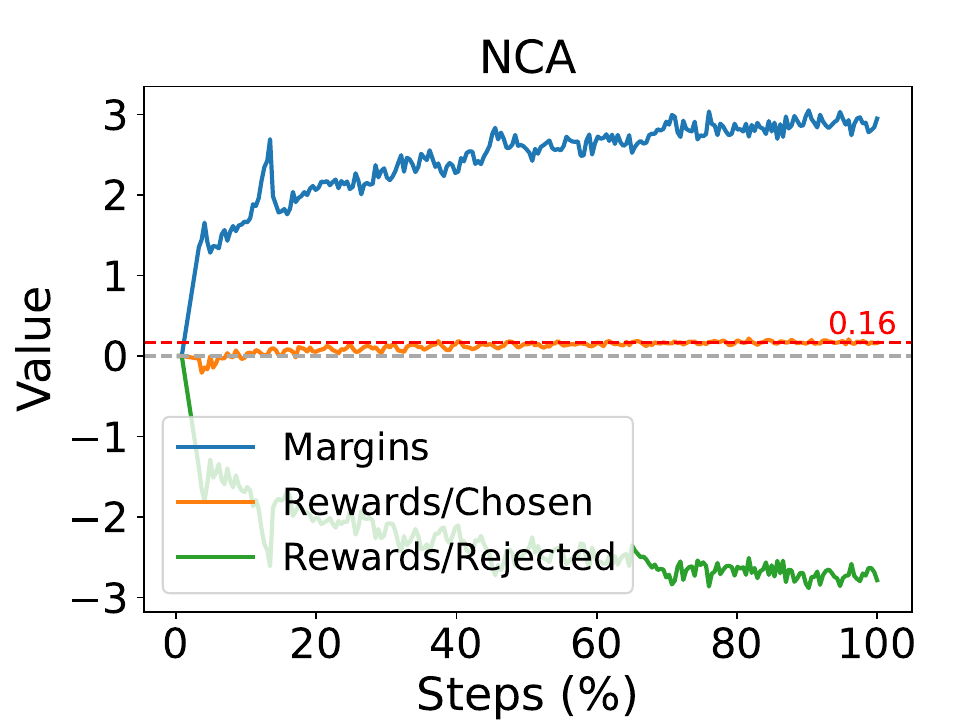}
    \vspace{-12pt}
    \caption{Reward patterns of \smallmodel preference learning with DPO, KTO, and NCA. For all algorithms, the rewards of rejected data keep decreasing and the margins between chosen and rejected data keep increasing. 
    However, the rewards of chosen data decrease below zero in DPO while keeping increasing and staying positive in KTO and NCA. 
    The absolute values of the reward in the last step (in red)
    of the three algorithms positively correlate with their performance in Table \ref{tab:sft}.}
    \label{fig:reward_pattern}
    \vspace{-5pt}
\end{figure}

\subsection{Explicit Reward as A Proxy? Hypothesis for Preference Learning in Reasoning}
\label{sec:reward_pattern}
We investigate the reason why DPO behaves differently than KTO and NCA. We start by empirically inspecting the rewards throughout the preference learning process, as shown in Figure \ref{fig:reward_pattern}. 
Rewards for chosen rejected data both keep decreasing through DPO, though the rewards for chosen data is still higher hence the loss decreases.
In KTO and NCA, the rewards of chosen data keep increasing with those of rejected data decreasing. 

\looseness=-1
Therefore, we hypothesize it is the distinction in the trend of rewards that leads to the performance gap between DPO and the other two algorithms. This distinction can be attributed to that DPO, derived from the Bradley-Terry model, only optimizes the relative differences between chosen and rejected data overlooking the absolute values of the rewards. 
This is a non-issue in alignment with general human values where preference is ``relative'' and there can be many valid answers to the same input.
However, in reasoning tasks, the space of correct answers is much smaller than that of incorrect ones.
Further, we notice that the rewards of chosen data in the last training step
follow the ranking order of KTO $>$ NCA $>$ DPO, positively correlate with their performance trends. 
Therefore, we believe that increasing the rewards of the chosen data is especially beneficial in preference learning for reasoning tasks.

\subsection{Ablation Study}
\input{tabs/ablation_study}

\looseness=-1
We study the impact of \dataname and other open-source alignment data on \smallsft's performance.
We consider three settings:
(1) \textbf{With original ground-truth answers,} which replaces the generated actions with ground-truth rationales and answers from the original datasets.
If no rationales are available, we use those from \dataname.
(2) \textbf{Open-source data only.}
(3)\textbf{\dataname only.}
We evaluate with the same setting as \S\ref{sec:exp} and report the averaged scores. See full results in Appendix \ref{sec:ablation-app}.

In Table \ref{tab:ablation}, \modelname outperforms the ``Grouth-truth'' model on all tasks, confirming the advantage of \dataname's designs of divide-and-conquer and code-as-action patterns, in line with conclusions of concurrent work \citep{Chen2024AgentFLANDD, Wang2024ExecutableCA}.
Training only on open-source data without \dataname greatly hurts the reasoning performance, confirming the effectiveness of \dataname. 
Meanwhile, training only on \dataname suffers a performance drop except for BBH, especially in instruction following. 
We attribute the performance drop to a worse instruction-following ability. This suggests the necessity of mixing \dataname with other alignment data for better all-around supervised fine-tuning.

%% file: tabs/ablation_study.tex
\begin{wraptable}{R}{0.5\textwidth}
\vspace{-23pt}
\caption{Ablation study of SFT data. 
}
\label{tab:ablation}
\resizebox{\linewidth}{!}{
\begin{tabular}{@{}lccccc@{}}
\toprule
\textbf{Model}                 & \textbf{Coding} & \textbf{Math} & \textbf{BBH} & \textbf{IFEval} & \textbf{Avg.} \\ \midrule
\textbf{\modelname-7B-SFT}    & 44.9            & 58.5          & 64.6         & 44.0            & 53.6          \\
\textbf{Ground-truth}       & 33.9            & 46.1          & 64.4         & 42.9            & 44.0          \\
\textbf{Open-source Only}   & 31.2            & 33.5          & 65.3         & 43.6            & 37.0          \\
\textbf{\dataname Only} & 37.3            & 56.2          & 67.0         & 17.4            & 47.7          \\ \bottomrule
\end{tabular}
}
\vspace{-10pt}
\end{wraptable}

%% file: sections/2_related_work.tex
\section{Related Work}

\textbf{Open LLMs in Reasoning.}
Open-source LLMs have shown remarkable progress in building \textit{specialists} that excel in mathematics reasoning~\citep{luo2023wizardmath, yue2023mammoth, toshniwal2024openmath} or coding abilities~\citep{roziere2023code,magicoder, deepseek-coder, zheng2024opencodeinterpreter}. 
On the contrary, mastering general reasoning capabilities still challenges open models, while the most advanced ones~\citep{deepseek-llm, qwen, Touvron2023Llama2O, Jiang2024MixtralOE} are well behind proprietary models. More, these cutting-edge open general-purpose models maintain their alignment recipes confidential, which further hinders the replication and development of open-source reasoning models.


\textbf{Preference Learning for Reasoning.}
Aligning language models from human or AI preferences has emerged as a prevalent approach in the open-source community ~\citep{tunstall2023zephyr, qwen} with the proposal of DPO~\citep{Rafailov2023DirectPO} and high-quality preference datasets~\citep{Cui2023UltraFeedbackBL, starling2023}. 
Different from open-domain chatbots, preference learning is largely underexplored in complex reasoning. Recent research showed performance degradation when applying DPO on reasoning tasks, but some newly proposed algorithms demonstrated a positive effect~\citep{Ethayarajh2024KTOMA, Chen2024NoiseCA, mitra2024orca, shao2024deepseekmath}. However, a deep understanding of preference learning, specifically its efficacy on complex reasoning, is not yet established.



%% file: sections/6_conclusion.tex
\section{Conclusion}
We strive to narrow the huge gap between open-source models and proprietary models from the perspective of alignment. Our work pushes the boundaries of open-source reasoning generalists by (1) releasing a high-quality multi-turn reasoning dataset \dataname with preference trees, (2) introducing \modelname-series LLMs which achieve new SOTA on challenging reasoning benchmarks and (3) providing insights on preference learning for reasoning through analysis, leading to new reward modeling objectives as well as a powerful reward model for reasoning. 

%% file: sections/7_appendix.tex
\appendix

\section{Additional Details in \dataname Construction}

\subsection{Dataset Details}
\label{sec:dataset_details}
\textbf{Math.} We adopt \textbf{GSM8K} \citep{cobbe2021training}, \textbf{MATH} \citep{hendrycks2021math}, \textbf{MathQA} \citep{amini-etal-2019-mathqa}, and \textbf{NumGLUE} \citep{mishra2022numglue}for mathematic reasoning, and include \textbf{TabMWP} \citep{lu2022dynamic} for tabular processing. 
We retain all the instructions for all datasets except MathQA, NumGLUE, and TabMWP.
MathQA divides problems into different categories according to the topics and annotates the formula that indicates the pattern needed to solve each problem. We apply stratified sampling to sample at most five problems for each pattern and prioritize the problems that come from the long-tail category.
Numglue contains eight different reasoning tasks and we discard Task 5 (Reading Comprehension + Explicit Numerical Reasoning), Task 6 (Reading Comprehension + Implicit Numerical Reasoning), and Task 7 (Quantitative NLI) due to the simplicity \cite{mishra2022numglue}.
For TabMWP, we only keep the questions with difficulty levels 4 and 5 since the rest are too easy for current state-of-the-art models. 

\textbf{Code.} We focus on programming with Python for the simplicity of integration of the interpreter. We use \textbf{CodeContest} \citep{li2022competition} and \textbf{TACO} \citep{Li2023TACOTI}, two competition-level coding datasets collected from various online platforms. We filter out the overlapped questions. Note that part of the questions in TACO only contain ground-truth solutions and do not contain test cases for evaluation, hence we apply GPT-4 to generate 12 test case inputs (4 basic inputs, 4 edge cases, and 4 large numbers) for each question and then execute the ground-truth solution snippets to produce outputs. Given that the two datasets mainly focus on competition problems that may deviate from real-world daily uses,  we exclusively adopt \textbf{Magicoder-Evol-Instruct} \citep{luo2023wizardcoder, magicoder}, the only dataset in our selection that does not contain test cases or ground-truth solutions. We employ GPT-4 Turbo to judge the correctness of generated code during interaction, and therefore we do not use this dataset for preference learning since we cannot rigorously construct pairs of correct and incorrect actions limited by the evaluation reliability.
We also include \textbf{WikiTableQuestions} \citep{pasupat2015compositional} for table processing with code.
 
\looseness=-1
\textbf{Logical Reasoning.} we use the multi-hop reasoning datasets \textbf{HotpotQA} \citep{yang2018hotpotqa} and \textbf{StrategyQA} \citep{geva2021strategyqa}, and the logical reasoning dataset \textbf{ReClor} \citep{Yu2020ReClorAR}. We follow the setting of \citet{Wang2023MINTEL} and convert HotpotQA to a generation task, removing the contexts and requiring LLMs to search relevant information using Wikipedia API.

\input{tabs/dataset_list}

\subsection{Details on Preference Tree Construction}
\label{sec:correct_action_generation_details}

\textbf{Models Adopted for Incorrect Action Sampling.}
We randomly sample one model from Mistral-7B-Instruct-v0.2, DeepSeek-Coder-33B-Instruct, Mixtral-8x7B-Instruct, and DeepSeek-LLM-67B-Chat to generate one incorrect action to pair with each correct one.

\textbf{Correct Action Generation Based on Ground Truth Annotations.}

We adopt GPT-3.5 Turbo as the generator to generate correct actions based on ground truth considering the instruction-following ability. We provide different access to the ground truth information for different tasks, specifically:
(1) For coding, where test cases are black boxes to reference solutions, we provide full access to the solution codes. The actor model will add step marks and corresponding explanations to the ground-truth code to make it easier to understand, or further refine the code for optimization. 
(2) For tool-free math problems, to avoid the actor model directly copying the answers to pass the correctness checking, we mask the answer numbers in the rationale before providing it to LLMs. This approach can better ensure response quality since it encourages LLMs to generate responses with complete reasoning chains with each step clearly marked.
(3) For program-enhanced math reasoning, we first translate the textual rationale into code. Then, we either directly provide it to the actor model to generate plans, or ask the actor model to convert the code into modularization programming and then make plans to create tools to solve problems.

\subsection{Data Decomtamination}
\label{sec:data_decomtamination}
We conduct careful decontamination. Firstly, for LeetCode, we apply the Exact Substring Matching Algorithm\footnote{\url{https://github.com/bigcode-project/bigcode-dataset/tree/main/decontamination}} to compare with each instruction in the \dataname and find no overlaps.
For others, we perform 8-gram exact matching to compare \dataname instructions with test sets of the same task.
We remove those instructions that overlap 8 grams with any test sample. 

\subsection{Detailed Statistics}
In total, \dataname has 86K instructions and 220K action pairs. 
The Total \# Pairs does not equal Total \# Turns in \dataname, since we fail to generate sufficient correct actions for every incorrect action in multi-turn trajectories mainly due to a lack of sufficient ground truth annotations. The total \# pairs may not equal \# correct answers, either, because it is also difficult and unnecessary to sample incorrect actions for the correct ones for some simple instructions.
\label{sec:stats_breakdown}
We present the specific information for each dataset. In particular, we list information on human annotation in each dataset, which plays an important role in correct action generation (\S \ref{sec:preference_data} and Appendix \ref{sec:correct_action_generation_details}).
All three steps of correct action sampling methods mentioned in \S \ref{sec:preference_data} can be applied to datasets that have rationales, while for datasets only containing answers, only the first two steps are applicable. We do not apply any of the three-step methods to generate correct answers for Magicoder, the only dataset without any human annotation, to construct preference pairs.

\input{tabs/stats_dataset_breakdown_new}

\section{Additional Details on Training \modelname Models}
\label{sec:training_details}
\textbf{Supervised Fine-Tuning.} 
 We finetune base models for 1 epoch with a 2e-5 learning rate and 0.1 warmup ratio using a cosine scheduler. For \modelname-7B, we mix 32K UltraChat, 30K ShareGPT, and 50K OpenOrca. For For \modelname-70B, we mix 63K UltraChat, 30K ShareGPT, and 70K OpenOrca.
 
\textbf{Preference Learning.} 
For hyperparameters, all $\beta$ is set to 0.1, and $\lambda_+$/$\lambda_-$ in KTO is set to 1.33 as recommended. We finetune models for 1 epoch with a 5e-7 learning rate and 0.1 warmup ratio using a cosine scheduler.

\textbf{Reward Modeling.}
 We train RM for 1 epoch with lr=1e-5 learning rate. We also use a cosine scheduler with a warmup ratio of 0.1. 

Regarding pair augmentation, we scale up the pairs by matching every correct action for each instruction with one incorrect action of other turns. This leads to NxN pairs of single-turn actions for a trajectory of depth N. We remove the action pairs consisting of nodes at the same turn, as they are already part of the multi-turn trajectory pairs we included.
Next, to avoid overfitting on the training set, we only select instructions with NxN $\leq$ 10, and for these instructions, we randomly sample at most 9 pairs with each action occurring no more than 3 times. This leads to an augmentation of 240k single-turn action pairs.

\input{tabs/additional_results}

\section{Additional Evaluation Results of \modelname}
\label{sec:additional_lm_eval}
\looseness=-1
\textbf{Detailed Setup in \S \ref{sec:exp}.}
For math, we test both textual reasoning and program-enhanced settings and report the best performance of the two. All evaluations are conducted in 0-shot CoT with two exceptions: BBH uses 3 shots and IFEval does not use CoT. 
For MINT, we select MATH, TheoremQA, and MMLU-math from ``reasoning'' as a new ``math'' split.
We also evaluate 5-shot MMLU \citep{hendrycks2021measuring} for STEM knowledge and MT-Bench \citep{Zheng2023JudgingLW} for conversation abilities to study whether \modelname needs to trade off other capabilities for reasoning.

\textbf{Results.}
Results are shown in Table \ref{tab:additional}.

On MMLU, \modelname outperforms baselines dedicated to coding and math, and achieves higher results than Mistral-Instruct-v0.2 and CodeLLaMA-70B-Instruct, the official aligned versions of our base model built by their authors.
Compared to general-purpose baseline models, \smallmodel achieves comparable performance with the top-performance OpenChat and Starling-LM, though \bigmodel does not achieve the same level of performance as other general-purpose models, which is expected due to the gap in the base models since CodeLLaMA-70B has not been intentionally optimized for knowledge.

On MT-Bench, we report baseline numbers from the official leaderboard\footnote{\url{https://huggingface.co/spaces/lmsys/chatbot-arena-leaderboard}}. \modelname matches the performance of mainstream open-source general-purpose models, and \bigmodel-KTO further achieves the score of GPT-3.5 Turbo.

\section{Detailed Results on Reward Modeling}
\label{sec:additional_rm_eval}
\subsection{Additional Results on Reranking}
\label{sec:bon}
\input{tabs/bon}
We present the full results on reranking in Table \ref{tab:best_of_n}, where the conclusions are consistent with those drawn from \S \ref{sec:additional_rm_eval}: 
(1) Our reward models always achieve the highest accuracy on all test sets across different N, except when N=2 on HumanEval.
(2) Both $\mathcal{L_{\text{BT}}}$ and $\mathcal{L_{\text{DR}}}$ consistently help improve reranking performance on three test sets except for HumanEval, where removing either of the objectives can prevent the accuracy from dropping when increasing N from 8 to 16.
(3) Modeling safety hurts reranking performance in reasoning. 
When removing UltraSafety from the training data, the RM achieves higher accuracies than \rewardmodel except on MBPP.


\section{Detailed Ablation Results}
\label{sec:ablation-app}
We present the full results of \S \ref{tab:ablation} in Table \ref{tab:ablation-app}, with detailed metrics on all coding and math datasets.
\begin{table}[t]
\caption{Ablation Study.}
\label{tab:ablation-app}
\resizebox{\textwidth}{!}{
\begin{tabular}{@{}l|ccc|ccccc|c|c|c@{}}
\toprule
\multirow{2}{*}{Model}         & \multicolumn{3}{c|}{Coding}                   & \multicolumn{5}{c|}{Math}                                                     & Reasoning     & Ins-Following & \multicolumn{1}{l}{\multirow{2}{*}{Avg.}} \\ \cmidrule(lr){2-11}
                               & HumanEval     & MBPP          & LeetCode      & GSM8K         & MATH          & TheoremQA     & SVAMP         & ASDiv         & BBH     & IFEval        & \multicolumn{1}{l}{}                      \\ \midrule
\textbf{\smallmodel-SFT}    & \textbf{55.5} & \textbf{59.1} & \textbf{20.0} & \textbf{73.7} & \textbf{32.6} & 20.0          & \textbf{82.2} & \textbf{84.1} & 64.6          & \textbf{44.0} & \textbf{53.6}                             \\
\textbf{Ground-Truth}       & 46.3          & 46.4          & 8.9           & 62.2          & 15.0          & 9.6           & 75.1          & 68.8          & 64.4          & 42.9          & 44.0                                      \\
\textbf{Open-Source Only}   & 38.4          & 44.1          & 11.1          & 45.3          & 10.8          & 9.3           & 52.7          & 49.4          & 65.3          & 43.6          & 37.0                                      \\
\textbf{\dataname Only} & 46.3          & 50.1          & 15.6             &  67.6         &   30.9        & \textbf{20.1} & 80.4          & 82.0          & \textbf{67.0} & 17.4          & 47.7                                      \\ \bottomrule
\end{tabular}
}
\end{table}

%% file: tabs/dataset_list.tex
\begin{table}[t]
\vspace{-10pt}
\caption{\dataname covers a diverse set of datasets spanning three tasks.}
\label{tab:data_list}
\small
\resizebox{\textwidth}{!}{

\begin{tabular}{@{}m{1.5cm} p{13cm}@{}}
\toprule
\bf Task   & \multicolumn{1}{c}{\bf Datasets}                                 \\ \midrule
\multirow{2}{*}{\bf Math}    & GSM8K \citep{cobbe2021training}, MATH \citep{hendrycks2021math}, MathQA \citep{amini-etal-2019-mathqa}, NumGlue \citep{mishra2022numglue}, TabMWP \citep{lu2022dynamic}    \\ \midrule
\multirow{2}{*}{\bf Coding} & CodeContest \citep{li2022competition}, TACO \citep{Li2023TACOTI}, WikiTableQuestions \citep{pasupat2015compositional}, Magicoder-Evol-Instruct \citep{luo2023wizardcoder, magicoder}\\ \midrule
{ \bf Logic}  & ReClor \citep{Yu2020ReClorAR}, HotpotQA \citep{yang2018hotpotqa}, StrategyQA \citep{geva2021strategyqa}           \\ \bottomrule
\end{tabular}
}
\end{table}

%% file: tabs/stats_dataset_breakdown_new.tex
\begin{table}[]
\caption{Stats breakdown}
\label{tab:stats_breakdown}
\resizebox{\textwidth}{!}{
\begin{tabular}{@{}l|cccccccc@{}}
\toprule
\multirow{2}{*}{\textbf{Task}} & \multirow{2}{*}{\textbf{Dataset}} & \multirow{2}{*}{\textbf{w/ Tool?}} & \multirow{2}{*}{\textbf{\# Prompts}} & \multirow{2}{*}{\textbf{\# Pairs}} & \multirow{2}{*}{\textbf{\# Correct Answers.}} & \multirow{2}{*}{\textbf{Avg. Length}} & \multicolumn{2}{c}{\textbf{Human Annotation}}           \\ \cmidrule(l){8-9} 
                               &                                   &                                    &                                      &                                    &                                               &                                       & \textbf{Has Answer?}      & \textbf{Has Rationale?}     \\ \midrule
\multirow{9}{*}{Math}          & \multirow{2}{*}{GSM8K}            & \Checkmark          & 4,522                                & 10,277                             & 17,392                                        & 1,746.7                               & \Checkmark & \Checkmark   \\
                               &                                   & \XSolidBrush        & 7,257                                & 10,879                             & 15,752                                        & 823.3                                 & \Checkmark & \Checkmark   \\
                               & \multirow{2}{*}{MATH}             & \Checkmark          & 7,474                                & 22,905                             & 34,667                                        & 1,189.0                               & \Checkmark & \Checkmark   \\
                               &                                   & \XSolidBrush        & 7,471                                & 25,765                             & 36,005                                        & 1,735.0                               & \Checkmark & \Checkmark   \\
                               & \multirow{2}{*}{MathQA}           & \Checkmark          & 7,552                                & 15,079                             & 20,328                                        & 2,338.5                               & \Checkmark & \Checkmark   \\
                               &                                   & \XSolidBrush        & 7,159                                & 17,743                             & 22,500                                        & 1,916.3                               & \Checkmark & \Checkmark   \\
                               & \multirow{2}{*}{NumGLUE}          & \Checkmark          & 3,020                                & 3,601                              & 5,717                                         & 1,474.6                               & \Checkmark & \XSolidBrush \\
                               &                                   & \XSolidBrush        & 2,835                                & 2,987                              & 4,273                                         & 1,056.1                               & \Checkmark & \XSolidBrush \\
                               & TabMWP                            & \Checkmark          & 3,117                                & 4,135                              & 6,083                                         & 842.6                                 & \Checkmark & \XSolidBrush \\ \midrule
\multirow{4}{*}{Coding}        & CodeContest                         & -                                  & 8,167                                & 44,319                             & 44,666                                        & 2,061.7                               & \Checkmark & \Checkmark   \\
                               & TACO                              & -                                  & 9,016                                & 50,877                             & 58,191                                        & 2,143.5                               & \Checkmark & \Checkmark   \\
                               & WikiTableQuestions                         & -                                  & 1,401                                & 1,544                              & 1,738                                         & 1,794.8                               & \Checkmark & \XSolidBrush \\
                               & Magicoder-Evol-Instruct                        & -                                  & 10,374                               & 0                                  & 10,238                                        & 687.1                                 & \XSolidBrush & \XSolidBrush \\ \midrule
\multirow{3}{*}{Logic}         & Reclor                            & \XSolidBrush        & 4,467                                & 7,958                              & 7,231                                         & 1,266.7                               & \Checkmark & \XSolidBrush \\
                               & HotpotQA                          & \Checkmark          & 1,182                                & 1,009                              & 1,230                                         & 1,333.2                               & \Checkmark & \XSolidBrush \\
                               & StrategyQA                        & \Checkmark          & 904                                  & 741                                & 968                                           & 1,256.2                               & \Checkmark & \XSolidBrush \\ \bottomrule
\end{tabular}
}
\end{table}

%% file: tabs/additional_results.tex
\begin{wraptable}{r}{0.4\textwidth}
\centering
\vspace{-25pt}
\caption{MMLU and MT-Bench.}
\label{tab:additional}
\resizebox{.4\textwidth}{!}{
\begin{tabular}{@{}lcc@{}}
\toprule
\multicolumn{1}{l|}{\textbf{Model}}                                 & \textbf{MMLU} & \textbf{MT-Bench} \\ \midrule
\multicolumn{3}{c}{$\sim$7B}                                                                                  \\ \midrule
\multicolumn{1}{l|}{\textbf{Mistral-7B-Instruct-v0.2}}                 & 58.9          & 7.60               \\
\multicolumn{1}{l|}{\textbf{Zephyr-7B-$\beta$}}                                & 59.7          & 7.34               \\
\multicolumn{1}{l|}{\textbf{OpenChat-3.5-1210}}                     & 63.4          & 7.81               \\
\multicolumn{1}{l|}{\textbf{Starling-LM-7B-$\alpha$}}                           & \textbf{64.0} & \textbf{8.09}      \\
\multicolumn{1}{l|}{\textbf{Magicoder-S-DS-6.7B}}                        & 37.1          & -                 \\
\multicolumn{1}{l|}{\textbf{OpenCI-DS-6.7B}}                        & 37.2          & -                 \\
\multicolumn{1}{l|}{\textbf{MAmmoTH-7B-Mistral}}                       & 56.2          & -                 \\
\multicolumn{1}{l|}{\textbf{WizardMath-7B-v1.1}}                       & 60.3          & -                 \\
\multicolumn{1}{l|}{\textbf{OpenMath-Mistral-7B}}                      & 58.3          & -                 \\
\rowcolor[HTML]{DAE8FC} 
\multicolumn{1}{l|}{\cellcolor[HTML]{DAE8FC}\textbf{\smallsft}}  & 61.8          & 7.15               \\
\rowcolor[HTML]{DAE8FC} 
\multicolumn{1}{l|}{\cellcolor[HTML]{DAE8FC}\textbf{+ DPO}}         & 62.4          & 7.38               \\
\rowcolor[HTML]{DAE8FC} 
\multicolumn{1}{l|}{\cellcolor[HTML]{DAE8FC}\textbf{+ KTO}}         & 62.2          & 7.38               \\
\rowcolor[HTML]{DAE8FC} 
\multicolumn{1}{l|}{\cellcolor[HTML]{DAE8FC}\textbf{+ NCA}}         & 62.2          & 7.38               \\ \midrule
\multicolumn{3}{c}{$\sim$40B}                                                                                 \\ \midrule
\multicolumn{1}{l|}{\textbf{Mixtral-8x7B-Instruct}}                 & \textbf{70.3} & \textbf{8.30}      \\
\multicolumn{1}{l|}{\textbf{DeepSeek-Coder-33B-Ins}}                    & 40.2          & -                 \\ \midrule
\multicolumn{3}{c}{$\sim$70B}                                                                                 \\ \midrule
\multicolumn{1}{l|}{\textbf{CodeLLaMA-70B-Instruct}}                & 55.1          & -                 \\
\multicolumn{1}{l|}{\textbf{DeepSeek-LM-67B-Chat}}                  & 72.3          & -                 \\
\multicolumn{1}{l|}{\textbf{QWen1.5-72B-Chat}}                          & \textbf{72.9} & \textbf{8.61}      \\
\multicolumn{1}{l|}{\textbf{OpenCI-CL-70B}}                         & 52.4          & -                 \\
\multicolumn{1}{l|}{\textbf{OpenMath-CL-70B}}                          & 60.2          & -                 \\
\rowcolor[HTML]{DAE8FC} 
\multicolumn{1}{l|}{\cellcolor[HTML]{DAE8FC}\textbf{\bigsft}} & 59.1          & 7.69               \\
\rowcolor[HTML]{DAE8FC} 
\multicolumn{1}{l|}{\cellcolor[HTML]{DAE8FC}\textbf{+ KTO}}         & 59.5          & 7.93               \\
\rowcolor[HTML]{DAE8FC} 
\multicolumn{1}{l|}{\cellcolor[HTML]{DAE8FC}\textbf{+ NCA}}         & 59.4          & 7.54               \\ \midrule
\multicolumn{3}{c}{Proprietary Models}                                                                  \\ \midrule
\multicolumn{1}{l|}{\textbf{GPT-3.5 Turbo}}                         & 70.0          & 7.94               \\
\multicolumn{1}{l|}{\textbf{GPT-4}}                                 & \textbf{86.4} & \textbf{8.96}      \\ \bottomrule
\end{tabular}
}
\vspace{-30pt}
\end{wraptable}

%% file: tabs/bon.tex
\begin{table}[tbh]
\caption{Detailed results of reranking Mistral-Instruct-v0.2's responses on coding and math.}
\label{tab:best_of_n}
\resizebox{\textwidth}{!}{
\begin{tabular}{@{}l|cccc|cccc|cccc|cccc@{}}
\toprule
Datasets                    & \multicolumn{4}{c|}{HumanEval}                                & \multicolumn{4}{c|}{MBPP}                                     & \multicolumn{4}{c|}{GSM8K}                                    & \multicolumn{4}{c}{MATH}                                      \\ \midrule
N                           & 2           & 4           & 8           & 16          & 2           & 4           & 8           & 16          & 2           & 4           & 8           & 16          & 2           & 4           & 8           & 16          \\ \midrule
Random                      & 41.5          & 39.0          & 40.2          & 39.6          & 33.1          & 33.6          & 34.3          & 30.1          & 45.0          & 43.1          & 44.5          & 40.2          & 11.5          & 11.3          & 10.0          & 8.5           \\
Top Logits                    & 43.3          & 43.3          & 43.3          & 43.3          & 35.3          & 35.3          & 35.3          & 35.3          & 45.7          & 45.7          & 45.7          & 45.7          & 12.1          & 12.1          & 12.1          & 12.1          \\
Self-Consistency            & 43.3          & 42.7          & 42.1          & 40.9          & 35.3          & 36.3          & 36.6          & 37.1          & 45.7          & 49.5          & 52.2          & 52.8          & 12.1          & 13.8          & 15.8          & 16.8          \\
Starling-RM-34B             & \textbf{47.6} & \textbf{47.0} & \textbf{49.4} & 45.7          & 37.8          & 38.8          & 39.6          & 40.4          & 49.1          & 52.8          & 56.0          & 56.5          & 6.5           & 7.2           & 7.7           & 7.7           \\ \midrule
\rewardmodel & 44.5          & 45.7          & 47.6          & 47.0          & 39.3          & \textbf{42.6} & \textbf{43.4} & \textbf{43.9} & \textbf{49.8} & 53.7          & 56.3          & 57.3          & 14.3          & 16.2          & 17.1          & 17.3          \\
w/o $\mathcal{L_{DR}}$      & 45.7          & 44.5          & 46.3          & 50.0          & 39.3          & 42.4          & 42.4          & 42.1          & 49.4          & 53.2          & 55.4          & 56.3          & 14.2          & 16.1          & 17.0          & 16.9          \\
w/o $\mathcal{L_{BT}}$      & 45.1          & 44.5          & 47.0          & 48.2          & 38.6          & 40.6          & 39.6          & 40.1          & 49.1          & 52.5          & 55.2          & 57.8          & 14.3          & 16.3          & 17.2          & 17.1          \\
w/o US                      & 45.7          & \textbf{47.0} & \textbf{49.4} & \textbf{50.6} & \textbf{39.3} & 41.1          & 41.4          & 42.9          & 49.4          & \textbf{53.8} & \textbf{57.4} & \textbf{58.7} & \textbf{14.5} & \textbf{16.6} & 17.2          & \textbf{17.5} \\
w/o UF + US                 & 43.9          & 43.3          & 47.0          & 46.3          & 36.3          & 38.1          & 36.6          & 35.3          & 49.4          & 52.3          & 54.6          & 57.2          & 14.3          & 16.5          & \textbf{17.4} & 17.4          \\ \midrule
Pass@N                      & 62.8          & 73.8          & 88.4          & 92.7          & 42.4          & 48.1          & 52.6          & 58.6          & 54.9          & 64.1          & 73.2          & 80.4          & 16.9          & 22.7          & 28.9          & 35.5          \\ \bottomrule
\end{tabular}
}
\end{table}